\documentclass[11pt]{article}
\usepackage[margin=1in]{geometry}
\usepackage{amsmath,amssymb}
\usepackage{graphicx}
\usepackage{booktabs}
\usepackage{tabularx}
\usepackage{xltabular}
\usepackage{caption}
\usepackage{placeins}

\usepackage{enumitem}
\usepackage{ragged2e}
\usepackage{microtype}
\usepackage{fontspec}
\usepackage{unicode-math}
\usepackage{newunicodechar}
\newunicodechar{−}{\ensuremath{-}}
\newunicodechar{–}{--}
\newunicodechar{—}{---}
\newunicodechar{γ}{\ensuremath{\gamma}}
\newunicodechar{β}{\ensuremath{\beta}}
\newunicodechar{κ}{\ensuremath{\kappa}}
\newunicodechar{π}{\ensuremath{\pi}}
\newunicodechar{ρ}{\ensuremath{\rho}}
\newunicodechar{α}{\ensuremath{\alpha}}
\newunicodechar{±}{\ensuremath{\pm}}
\newunicodechar{≤}{\ensuremath{\le}}
\newunicodechar{∈}{\ensuremath{\in}}
\newunicodechar{ε}{\ensuremath{\epsilon}}
\newunicodechar{×}{\ensuremath{\times}}
\newunicodechar{Δ}{\ensuremath{\Delta}}
\newunicodechar{₀}{\textsubscript{0}}
\newunicodechar{₁}{\textsubscript{1}}
\newunicodechar{₂}{\textsubscript{2}}
\newunicodechar{≈}{\ensuremath{\approx}}
\newunicodechar{≠}{\ensuremath{\ne}}
\newunicodechar{⁴}{\textsuperscript{4}}
\newunicodechar{→}{\ensuremath{\rightarrow}}
\newunicodechar{·}{\ensuremath{\cdot}}
\newunicodechar{½}{\textonehalf}
\newunicodechar{†}{\textdagger}
\newunicodechar{‡}{\textdaggerdbl}
\newunicodechar{⁻}{\textsuperscript{-}}
\newunicodechar{≥}{\ensuremath{\ge}}
\newunicodechar{°}{\textdegree}
\newunicodechar{∎}{\ensuremath{\blacksquare}}
\newunicodechar{∪}{\ensuremath{\cup}}
\usepackage[colorlinks=true,linkcolor=blue,citecolor=blue,urlcolor=blue]{hyperref}

\title{Beyond Balanced Accuracy: A Resolution- and Parity-Controlled Benchmark for Vision–Language and Vision-Only Defect Assessment in UAV Power-Line Inspection}
\author{Linghao Zhang\textsuperscript{1,2}, Siyu Xiang\textsuperscript{1,2}, Junwei Kuang\textsuperscript{1,2}, Peiyu Yi\textsuperscript{1,2}\\[1ex]
\textsuperscript{1}State Grid Sichuan Electric Power Research Institute, Chengdu 610041, China\\
\textsuperscript{2}Power System Security and Operation Key Laboratory of Sichuan Province\\[1ex]
\texttt{16100178@qq.com}, \texttt{xiangsiyu4024@163.com}\\
\texttt{kuangjunwei1258@163.com}, \texttt{yipeiyu1994@gmail.com}}
\date{September 21, 2026}

\begin{document}
\maketitle

\begin{abstract}
Vision–language models (VLMs) are often reported to outperform task-specific vision backbones for unmanned aerial vehicle (UAV) power-line defect assessment. We test that claim on ElecVQA-Bench, a 56,972-item benchmark derived from the public InsPLAD dataset, across six evaluation choices: partition, evaluated item set, label space, replication, input resolution, and side information. On a matched partition, a Swin Transformer and the strongest adapted VLM differ by only 0.03 points at binary screening. At seven-way defect typing, increasing the vision backbones from 224 px to the measured pixel budget of the VLM preprocessor narrows the gap against InternVL3.5-8B from +20.53 to −0.57 points for ResNet-50 and from +23.67 to +4.70 points for Swin-T. A pixel-budget audit shifts Qwen3-VL-8B macro recall by 10.78 points, yet a source-pixel-matched InternVL control still leaves Qwen ahead by 7.43 to 13.61 points while using 56\% fewer visual tokens, so neither source pixels nor token budget explains the difference between the two VLMs. A two-seed global replication changes Qwen binary accuracy and seven-way macro recall by 0.86 and 1.02 points. After split-specific retraining, Qwen does not lead at crop or image level, and a 14-tower, three-seed replication reverses the sign across seeds, giving mean common-six macro recall of 0.9085 for Qwen against 0.9509 for ResNet-50. No split regime yields a family-level advantage that survives multiple-comparison correction. The study supports a benchmark-audit contribution rather than a general claim of VLM superiority.
\end{abstract}


\vspace{0.5em}

\section*{1. Introduction}
Inspection of overhead transmission infrastructure is critical to grid reliability, and UAV imagery has become the dominant acquisition mode. Two model families compete to interpret that imagery. Purpose-trained vision backbones—convolutional networks and, increasingly, vision transformers—are the incumbent option: compact, fast, and well understood. Vision–language models are the challenger, motivated by the argument that web-scale pre-training yields better few-shot behaviour and an interpretable rationale rather than a bare class label.

We do not attempt to settle that debate here. Our claim is narrower: the reported outcome depends on six experimental choices that are easy to get wrong and are often left unreported, and once those choices are controlled, no stable family-level ordering remains on this benchmark.

The first choice is the evaluation partition. An initial comparison scored the two families on different test files: a seven-class file of 7,851 items for the vision backbones and a binary file of 2,021 items for the VLMs. That mismatch produced a VLM lead of 13 to 15 points. Re-scoring the vision backbones on the identical 2,021-item file, with nothing else changed, produced no detectable difference.

The second choice is input resolution, which to our knowledge no comparison in this literature reports for the VLM side. The VLM preprocessors tile images at 448 px. An audited inference pass records a mean of 2.94 patches per T3 item for InternVL3.5-8B, an aggregate pixel budget equivalent to a 768 px square image, against the 224 × 224 input conventionally used for vision backbones. At seven-way typing, this difference accounts for most of the apparent VLM advantage: moving each backbone to the two resolutions nearest that budget changes the gap against InternVL3.5-8B from +20.53 to −0.57 points for ResNet-50 and from +23.67 to +4.70 points for Swin-T. It does not close the gap against Qwen3-VL-8B, and a source-pixel-matched control does not close it either.

The third choice is replication. A 14-tower, three-seed replication reverses the sign of the tower-level comparison across seeds: ResNet-50 leads in two of three matched seeds, with a seed standard deviation of 0.5 points against 6.2 points for Qwen. Split construction and seed therefore shape the direction of the result at least as strongly as model family does.

The fourth and fifth choices are the label space and the evaluated item set, both of which are shown below to move the comparison by more than ten points when left uncontrolled. The sixth is side information. The VLM receives an asset name, a defect decision tree, and a per-asset candidate option list, while the vision-only classifier receives only pixels and a fixed label set. Every cross-family comparison in this paper is therefore a system-level comparison, and we label it as such. Abstention is closely related: the benchmark offers an explicit "cannot judge" option, zero-shot models use it heavily, and one model leaves 20\% of defective items unjudged, so a risk function that prices abstention at zero systematically rewards that behaviour.

Our contributions are as follows.

\begin{enumerate}
\item A parity-audit protocol covering test file, evaluated item set, label space, replication count, input resolution, and side information, with an explicit verdict attached to every cross-family comparison reported here.
\item A resolution, leakage, and retraining audit. Moving each vision backbone from 224 px to the resolutions nearest the measured pixel budget changes the gap against InternVL3.5-8B from +20.53 to −0.57 points for ResNet-50 and from +23.67 to +4.70 points for Swin-T. A pixel-budget audit changes 4,472 of 7,284 Qwen inputs and moves Qwen macro recall by 10.78 points, while a source-pixel-matched InternVL control changes 4,699 inputs and leaves Qwen ahead at every setting. After crop and image retraining, tower-clustered testing, and a 14-tower three-seed replication, no split regime retains a stable Qwen–ResNet advantage.
\item Abstention treated as a reported evaluation quantity. We show that balanced accuracy cannot separate error profiles dominated by misses, by hallucinations, or by abstentions, extend the cost-sensitive risk function to price refusal, and document that the model selected as risk-minimizing under a zero-price analysis leaves 20\% of defective items unjudged while reporting a zero miss rate.
\item A prevalence-conditional reading of cost-sensitive rankings. The crossover cost ratio scales with the odds of the negative class, so a ratio derived on a constructed benchmark does not transfer to a field prevalence two orders of magnitude lower. We give the closed-form re-basing and the resulting surface.
\item A decomposition showing that macro-averaging over a long-tailed label set dominates the headline metric. Two classes holding nine test instances between them move the seven-class macro-average by 22.9 points in opposite directions.
\item ElecVQA-Bench, a benchmark of 56,972 items built from the public InsPLAD dataset \cite{ref1} by a deterministic pipeline and released with construction code, split manifests, prompt templates, evaluation harness, and adapter weights.
\end{enumerate}

The remainder of the paper is organized as follows. Section 2 reviews related work on power-line inspection, cost-sensitive evaluation, and vision–language modelling. Section 3 describes the construction and statistics of ElecVQA-Bench. Section 4 presents the cost-sensitive framework, the parity audit, and the experimental configuration. Section 5 reports results. Section 6 discusses what the controlled comparisons establish, states the limitations of the study, and outlines future work. Section 7 concludes.

\section*{2. Related Work}
Power-line inspection from UAV imagery. InsPLAD \cite{ref1} provides 10,607 UAV images across 17 asset categories with attached defect-classification and anomaly-detection subsets, together with baselines using DifferNet \cite{ref2} and CS-Flow \cite{ref3}, both normalizing-flow density estimators originally proposed for industrial surface inspection. Nguyen et al. \cite{ref4} survey vision-based power-line inspection and identify data scarcity and class imbalance as recurring obstacles. Tao et al. \cite{ref5} detect insulator defects from aerial imagery with a cascaded convolutional architecture, and the wider literature builds predominantly on general-purpose detectors such as Faster R-CNN \cite{ref6} and the YOLO family \cite{ref7}. These works report detection or classification accuracy on their own partitions. We did not identify, in the literature reviewed here, a study that verifies partition and label-space equivalence before comparing model families. This review is selective rather than systematic, and motivates the audit rather than claiming exhaustive coverage.

Industrial defect and anomaly detection. The evaluation conventions adopted here—crop-level scoring, heavy normal-to-defect imbalance, and a small number of rare defect categories—follow the industrial anomaly-detection literature established by MVTec AD \cite{ref8} and memory-bank methods such as PatchCore \cite{ref9}. Recent surveys \cite{ref10} document the migration of that field toward foundation-model backbones; we examine the evaluation methodology behind that migration.

Cost-sensitive evaluation. That aggregate accuracy is an unsafe summary under asymmetric misclassification costs is long established. Provost and Fawcett \cite{ref11} argue for evaluating classifiers across the full range of plausible cost ratios rather than at a single operating point; Drummond and Holte \cite{ref12} formalize this as cost curves; Elkan \cite{ref13} gives the general reduction of cost-sensitive learning to a threshold problem. Calibration research \cite{ref14} separately shows that aggregate correctness does not certify that a model's confidence, or in a discrete answer space its qualitative error profile, is trustworthy. We adapt this reasoning to the miss, hallucination, and abstention vocabulary of industrial visual question answering.

Class imbalance and long-tailed recognition. The seven-way results reported here are macro-averaged over a label set containing a class with four test instances, which places them in the long-tailed regime analysed by Buda et al. \cite{ref15}. Class-balanced reweighting \cite{ref16} and the representation–classifier decoupling of Kang et al. \cite{ref17} are the standard mitigations. None is applied here, and Section 5.7.2 quantifies the consequences of that choice for the headline comparison.

Vision–language models and hallucination evaluation. General-purpose VLMs derive from contrastive pre-training \cite{ref18} and instruction tuning \cite{ref19,ref20}. InternVL \cite{ref21} and Qwen2.5-VL \cite{ref22} are used here as representative open families. Because the experimental focus is evaluation methodology rather than architecture, we do not summarize the technical reports of every model generation. The hallucination rate defined in Section 4.1 is the industrial-inspection analogue of the object-hallucination metrics of CHAIR \cite{ref23} and POPE \cite{ref24}, with the difference that the present setting has a closed answer space and an explicit abstention token, so hallucination and refusal are separately measurable rather than confounded.

Parameter-efficient adaptation and statistical testing. LoRA \cite{ref25} and its quantization-aware variant QLoRA \cite{ref26} are the default route to adapting large models on a single accelerator. The multi-task ablation is grounded in the classical result that jointly trained tasks can regularize one another \cite{ref27}, tempered by the observation that task pairing depends on task relatedness \cite{ref28}. ResNet-50 \cite{ref29} is used as the convolutional baseline and Swin-T \cite{ref30} as the hierarchical vision-transformer baseline. McNemar's exact test \cite{ref31} is applied following the analysis of paired tests for classifier comparison by Dietterich \cite{ref32}.

\section*{3. Materials: ElecVQA-Bench}
\subsection*{3.1. Source Data and Construction}
ElecVQA-Bench is built from three InsPLAD subsets. The public release describes 10,607 UAV images across 17 source asset categories. After the pre-construction audit, the detection subset retained here contains 10,552 images, 28,824 COCO boxes, and 18 normalized asset categories. The supervised fault-classification subset contains 11,525 crops over 5 asset types and 6 defect categories, and the unsupervised anomaly-detection subset contains 26,825 crops over 17 asset types with 695 anomalous items. The raw archive is read-only and every derived artefact is reconstructible from released scripts. The pre-construction audit recorded category-naming inconsistencies, categories with zero validation instances, degenerate bounding boxes, crop-size distributions, and near-duplicate frames.

Algorithm 1 gives the three-stage construction procedure: region-of-interest extraction at four expansion ratios, question construction for three tasks, and a leakage-aware, parent-image-grouped split with automated assertions.

\begin{quote}
Algorithm 1. ElecVQA-Bench construction
Input: raw InsPLAD subsets; expansion ratios R = \{1.0, 1.5, 2.0, 3.0\}
Output: item sets for T1 to T3; digested split manifest
\begin{enumerate}
\item Audit source data for category mapping, duplication, and box validity.
\item For each annotated instance, expand its box by every r ∈ R about its centre, clip to the image, and enlarge to a 64 px minimum short side.
\item T1 (grounding): image, asset name, normalized box.
\item T2 (binary assessment, r = 2.0): crop, P3 prompt, label ∈ \{normal, defect\}. "Cannot judge" is an available answer but never a ground-truth label. Negatives are drawn from \texttt{good} crops of the same asset type.
\item T3 (fine-grained typing, r = 2.0): crop, P3 prompt, label ∈ closed defect set ∪ \{good\}. "Cannot judge" is again available but not a ground-truth label.
\item Group items by the filename-derived \texttt{parent\_id}. Apply a group-level 70/10/20 split with seed 42, stratified by each group's dominant defect class. Asset–label combinations with fewer than five instances are assigned to the test partition.
\item Assert that no \texttt{parent\_id} crosses a split, that no perceptual-hash collision crosses a split, and that every class present in the training split has at least five training instances. Test-only rare classes are reported separately and are not covered by the training-support assertion.
\item Write SHA-256 digests of splits, scripts, and prompts to a manifest.
\end{enumerate}

\end{quote}

Two design choices matter for the safety metrics of Section 4.1. Questions are phrased neutrally, for example "judge the condition of the glass insulator shown" rather than "what defect does this insulator have", and an abstention option is always present, so refusal is measurable rather than unparseable. Because no training item carries abstention as a target, a high abstention rate reflects a model's disposition under the prompt rather than a data-driven uncertainty estimate.

\subsection*{3.2. Benchmark Statistics}
Table 1 gives the item counts. Split proportions are the realized output of the \texttt{parent\_id}-grouped split described in Algorithm 1. Because group sizes are uneven, realized proportions deviate from the 70/10/20 target by up to 1.5 percentage points, and realized counts rather than target ratios are authoritative. The identifier \texttt{parent\_id} is filename-derived and is not a globally normalized source-photograph identifier; the source-photograph overlap audit is reported in Section 6.4.

\begin{table}[htbp]
\centering
\setlength{\tabcolsep}{3.0pt}
\renewcommand{\arraystretch}{1.15}
\caption*{Table 1. ElecVQA-Bench composition.}
\begin{tabularx}{0.645\textwidth}{>{\hsize=2.131\hsize\RaggedRight\hyphenpenalty=10000\exhyphenpenalty=50\arraybackslash}X >{\hsize=0.627\hsize\RaggedRight\hyphenpenalty=10000\exhyphenpenalty=50\arraybackslash}X >{\hsize=0.627\hsize\RaggedRight\hyphenpenalty=10000\exhyphenpenalty=50\arraybackslash}X >{\hsize=0.988\hsize\RaggedRight\hyphenpenalty=10000\exhyphenpenalty=50\arraybackslash}X >{\hsize=0.627\hsize\RaggedRight\hyphenpenalty=10000\exhyphenpenalty=50\arraybackslash}X}
\toprule
Task & Total & Train & Validation & Test \\
\midrule
T1, grounding & 10,605 & 7,448 & 1,068 & 2,089 \\
T2, binary assessment & 9,997 & 6,949 & 1,027 & 2,021 \\
T3, fine-grained typing & 36,370 & 25,460 & 3,626 & 7,284 \\
Total & 56,972 & 39,857 & 5,721 & 11,394 \\
\bottomrule
\end{tabularx}
\end{table}

T3's 36,370 items correspond to one item per crop at r = 2.0 over the union of the fault-classification and anomaly-detection subsets, that is 38,350 candidate crops less 1,980 (5.2\%) removed by the Algorithm 1 audit. The per-reason breakdown is released as \texttt{construction\allowbreak\_audit.csv}. T1 is constructed and released for future grounding experiments but is not evaluated in this paper.

T2's test set contains 1,255 normal and 766 defective items, giving π\textsubscript{N} = 0.621 and π\textsubscript{D} = 0.379. This is not the 1:1 ratio targeted at construction. Negatives are matched 1:1 to positives before splitting, but grouping by \texttt{parent\_id} distributes them unevenly, and defective crops cluster more strongly by parent group than \texttt{good} crops do. The realized prevalence of 37.9\% is therefore a property of this split rather than a designed value, and Section 4.4 shows it is the most influential quantity in the cost analysis.

The T3 test set covers six defect types, namely \texttt{rust}, \texttt{corrosão}, \texttt{missing-cap}, \texttt{nest}, \texttt{torned-up}, and \texttt{peeling-paint}, plus \texttt{good}, giving K = 7 ground-truth classes. T2 uses the binary label \texttt{normal} for the no-defect class and T3 preserves the source label \texttt{good} for the same concept; Table A3 gives the crosswalk. All T3 balanced accuracies reported here are unweighted macro-averaged recall over exactly these seven classes, with abstentions scored as errors against the true class. The label \texttt{corrosão} retains the original Portuguese term used by InsPLAD and denotes generalized corrosion, whereas \texttt{rust} denotes localized oxidation on metal fittings. Per-class test support is given in Table A4. Its most consequential entries are \texttt{peeling-paint}, a test-only class with four test instances, and \texttt{torned-up}, with five test instances; neither figure is a whole-archive count.

\subsection*{3.3. Prompts and Evaluation Protocol}
Four graded prompt levels are released. P1 contains the question stem and the candidate answer options. P2 adds one-sentence textual defect definitions. P3 replaces those definitions with an ordered decision tree. P4 adds one reference image per class. A P3 prompt therefore has the fixed structure: asset name, asset description, decision tree, question, candidate answer options. All main experiments fix P3 for both training and inference, with the prompt version recorded in every run identifier. Outputs are constrained to a JSON schema by guided decoding, a 20-item smoke test precedes every full run, and format-failure rate is reported separately from accuracy. It was 0.0\% in every run. Exact prompt templates and per-asset option lists are released with the benchmark manifest.

We declare three asymmetries between the two families and carry them through the analysis. First, guided decoding guarantees that the VLM emits a syntactically legal label, whereas a vision-only classifier emits one by construction, so the VLM solves a constrained generation problem and the classifier a discriminative one. Second, the VLM receives only the answer options possible for the asset in question, typically two to four, while the classifier performs unconstrained seven-way classification and can in principle predict an impossible class. Analysis of the stored predictions shows this affects 7 of 7,284 items (0.1\%), and oracle correction of those items changes balanced accuracy by at most 0.55 points. Third, the VLM prompt supplies textual side information in the form of an asset name and a decision tree, while the classifier receives only the image and the fixed label set. Cross-family comparisons are therefore system-level comparisons, and no result in this paper claims that the two arms receive equivalent input information.

\section*{4. Methods}
\subsection*{4.1. Notation}
For T2, let Y ∈ \{N, D\} be the ground truth and Ŷ ∈ \{N, D, A\} the model output, corresponding to normal, defective, and abstain. Define the miss rate β = P(Ŷ = N | Y = D) and the hallucination rate α = P(Ŷ = D | Y = N), with class-conditional abstention rates γ\textsubscript{D} = P(Ŷ = A | Y = D) and γ\textsubscript{N} = P(Ŷ = A | Y = N). Then TPR = 1 − β − γ\textsubscript{D}, TNR = 1 − α − γ\textsubscript{N}, and

\[\mathrm{BalAcc}=\frac{1}{2}(\mathrm{TPR}+\mathrm{TNR})=1-\frac{1}{2}(\alpha+\beta+\gamma_D+\gamma_N).\]
Equation (1) is an identity rather than an approximation. Section 5.1 uses it to cross-validate the measured abstention rates.

\subsection*{4.2. The Balanced-Accuracy Blind Spot}
Proposition 1. Fix BalAcc = b and write s = 2(1 − b). By Equation (1) the set of error profiles consistent with b is

\[\Delta_b=\{(\alpha,\beta,\gamma_D,\gamma_N)\in[0,1]^4:\alpha+\beta+\gamma_D+\gamma_N=s,\ \alpha+\gamma_N\le1,\ \beta+\gamma_D\le1\}.\]
a three-dimensional polytope. Consequently, for any 1/2 ≤ b < 1 there exist models of identical balanced accuracy whose errors are dominated respectively by misses (β → s), by hallucinations (α → s), and by abstentions (γ\textsubscript{D} + γ\textsubscript{N} → s), and no function of b alone distinguishes them.

Proof. Equation (1) imposes one linear equality on four non-negative quantities subject to two row-stochasticity constraints. The three named cases are the coordinate extrema of Δ\textsubscript{b} and are feasible whenever s ≤ 1.

The argument is deliberately elementary. The degeneracy is a property of any symmetric two-class summary and underlies the classical case for cost curves over single-number accuracy \cite{ref11,ref12}. Section 5.1 shows that all three extreme cases are occupied by real models in this study, and that the abstention-dominated case is the one an accuracy-only report renders invisible.

\subsection*{4.3. Cost-Sensitive Risk with Abstention}
Let C\textsubscript{m} be the cost of a missed defect, C\textsubscript{h} the cost of a false alarm, and C\textsubscript{a} the cost of an abstention, which in an inspection workflow is the cost of routing the item to manual review. Setting C\textsubscript{h} = 1 without loss of generality and writing ρ = C\textsubscript{m}/C\textsubscript{h} and κ = C\textsubscript{a}/C\textsubscript{h}, the expected cost per item under model i is

\[R_i(\rho,\kappa)=\pi_D(\rho\beta_i+\kappa\gamma_{D,i})+\pi_N(\alpha_i+\kappa\gamma_{N,i}).\]
The case κ = 0 assigns zero cost to a refusal. A model can therefore lower its apparent risk without limit by abstaining. We restrict attention to 0 ≤ κ ≤ ρ as an economically interpretable range in which manual review is no more costly than the failure it prevents. This assumption concerns the inspection setting; it is not a universal claim. The cases examined here are bracketed by κ = 0 and κ = 1. For two models A and B with β\textsubscript{A} ≠ β\textsubscript{B}, equating risks gives the crossover cost ratio

\[\rho^*(\kappa)=\frac{\pi_N(\alpha_B-\alpha_A)+\kappa[\pi_D(\gamma_{D,B}-\gamma_{D,A})+\pi_N(\gamma_{N,B}-\gamma_{N,A})]}{\pi_D(\beta_A-\beta_B)}.\]
\begin{quote}
Algorithm 2. Abstention-aware cost-ratio crossover
Input: (α, β, γ\textsubscript{D}, γ\textsubscript{N}) for models A and B; priors π\textsubscript{D}, π\textsubscript{N}; abstention cost κ ≥ 0
Output: ρ* and the risk-preferred model on each side, or a dominance verdict
\begin{enumerate}
\item If β\textsubscript{A} = β\textsubscript{B}, there is no crossover in ρ. Compare R\textsubscript{A} and R\textsubscript{B} directly; the lower value dominates for all ρ.
\item Otherwise compute ρ*(κ) from Equation (3).
\item If ρ* ≤ 0, the crossover lies outside the physical domain. Report global dominance of whichever model has lower risk at ρ = 0 and do not present ρ* as a decision threshold.
\item If 0 < ρ* < 1, the crossover implies that a missed defect is cheaper than a false alarm. Report it but flag it as outside the regime the framework addresses.
\item If ρ* ≥ 1, evaluate R\textsubscript{A}(1, κ) and R\textsubscript{B}(1, κ) to identify the preferred model below ρ*; the other is preferred above.
\item Repeat over the plausible ranges of κ and π\textsubscript{D} and report ρ* as a surface rather than a point.
\end{enumerate}

\end{quote}

Because the models compared here are discrete classifiers with no tunable threshold, and because their output is a token rather than a score, a full cost curve in the sense of Drummond and Holte \cite{ref12} is unavailable without a confidence signal. This study does not extract such a signal. The abstention term is related to reject-option and selective-prediction frameworks and is used here as an accounting device rather than as a new theory of selective classification.

\subsection*{4.4. Prevalence Sensitivity}
Equation (3) is proportional to π\textsubscript{N}/π\textsubscript{D}. The benchmark's π\textsubscript{D} = 0.379 is a construction artefact rather than an estimate of prevalence on a real line, where reported defect rates are one to two orders of magnitude lower. Writing ρ*₀ for the value at π\textsubscript{D} = π₀, the value at any other prevalence π follows without re-running any model:

\[\rho^*(\pi)=\rho^*_0\cdot\frac{\pi_0}{1-\pi_0}\cdot\frac{1-\pi}{\pi},\quad \kappa=0.\]
Equation (4) applies to the zero-cost abstention case; the general surface is given by Equation (3). Any crossover ratio derived from a constructed benchmark therefore carries an implicit prevalence assumption that must be reported with it.

\subsection*{4.5. Parity Audit}
\begin{quote}
Algorithm 3. Parity audit
Input: result records R₁ and R₂, each with test file, evaluated item-identifier list, item count n, ground-truth class set C, seed count, and side-information set S
Output: PASS, PASS with caveat, or FAIL with the specific discrepancy
\begin{enumerate}
\item If test files differ: FAIL ("different files").
\item If n₁ ≠ n₂: FAIL ("different item counts").
\item If C₁ ≠ C₂: FAIL ("different label spaces").
\item If evaluated item-identifier sets differ despite equal n: FAIL ("same size, different items"). This subsumes the case in which one record is scored on a subsample of the other's file.
\item If seed counts differ, or either is below three: PASS with caveat. Any difference smaller than the wider seed-to-seed spread must not be reported as a result.
\item If S₁ ≠ S₂: PASS with caveat ("system-level comparison only"). Such a result may be reported as a whole-system comparison but not as evidence that the two families received equivalent input information.
\item Otherwise: PASS.
\end{enumerate}

\end{quote}

Lines 4 to 6 are the substantive checks. Line 1 alone is insufficient when one record reports a subsample of the other's file; line 5 prevents a difference smaller than the observed run-to-run dispersion from being reported as a finding; and line 6 prevents a system-level difference from being read as input parity. Every cross-family comparison in Section 5 carries its audit verdict, and every such comparison fails line 6 by construction.

\subsection*{4.6. Models, Adaptation, and Statistical Testing}
The contrast throughout is between language-grounded generative models (VLMs) and vision-only discriminative backbones, the latter comprising one convolutional network, ResNet-50 \cite{ref29}, and one hierarchical vision transformer, Swin-T \cite{ref30}. Because Swin-T is itself a transformer and appears on the losing side of several vision-only comparisons, the coarse convolutional-versus-transformer explanation is excluded. Architecture-specific differences in visual encoding are not excluded and are treated in Section 5.7.4 as an open candidate.

Seven models were evaluated zero-shot: InternVL3.5-2B and -8B, Qwen3-VL-2B and -8B, and Qwen2.5-VL-7B with open weights, plus Qwen3.8-27B and Qwen3.5-122B-A10B-FP8 through a hosted API with \texttt{enable\_thinking = false}. Adaptation used LoRA \cite{ref25} at rank 32 with α = 64 and dropout 0.05 on the target modules listed in Table A5, with the vision encoder frozen, a learning rate of 10⁻⁴ with a cosine schedule and 3\% warmup, three epochs unless stated otherwise, an effective batch of 16, and bf16 precision, training approximately 0.25\% to 0.41\% of parameters. The vision-only baselines were ImageNet-initialized and trained for 10 epochs with the augmentation and normalization settings listed in Table A5, at 224 × 224 unless a resolution is stated.

Two scoring granularities are used on T3. The seven-way score is macro-averaged recall over the seven ground-truth classes, obtained by mapping each model's answer-option output back to those labels. The three-way score is macro-averaged over the coarser option-level partition and is not a set-theoretic collapse of the seven-way label space. The two average over different partitions with different weights, so neither is constrained to exceed the other and no monotonic relation between them should be inferred.

Input resolution is treated as a controlled variable. The VLM preprocessors tile at 448 × 448 with dynamic tiling to a maximum of 12 tiles and 256 visual tokens per tile. An audited inference pass records, for InternVL3.5-8B on T3, a mean of 2.94 patches and 752.9 visual tokens per item; 2.94 tiles of 448 × 448 is an aggregate pixel budget equivalent to a 768 px square image. The vision backbones were re-trained at 224, 448, 672, and 896 px, so the matched budget lies between the 672 px and 896 px settings rather than coinciding with either. Qwen3-VL-8B's processor behaves differently, recording a mean of 329.5 and a median of 144 visual tokens per T3 item at native settings, so the two VLMs are not token-matched even when their source-pixel budgets are. The pixel-budget experiments of Section 5.7.1 were run through an instrumented inference path that additionally logs grids, patch and token counts, input hashes, and prediction hashes; its native-setting macro recalls differ from the main adaptation pass by 0.31 points for Qwen3-VL-8B and 0.56 points for InternVL3.5-8B. All comparisons are made within a single pass.

{\emergencystretch=2em
Quantization used bitsandbytes NF4 and INT8 weight-only with bf16 activations, scored through the identical harness on the identical items as the bf16 reference. Where a proportion is observed to be zero, we report the one-sided 95\% upper bound 3/n rather than 0.00\%.\par}

For item-level contrasts on a shared evaluated item set we use McNemar's exact test \cite{ref31,ref32} on paired per-item correctness, with Holm correction within each stated family of comparisons. For the primary macro-recall contrasts, uncertainty and significance are computed at the level of the sampling cluster rather than the individual crop, using cluster-level sign-flip randomization with 10,000 replicates. In each replicate all predictions within a cluster are either kept or swapped between the two models, which preserves within-cluster dependence and the class distribution. The unshifted null distribution gives the two-sided p-value, and the null distribution shifted by the observed difference gives the 95\% randomization interval, so the interval and the test are derived from the same reference distribution and cannot disagree. Class support is preserved in every replicate, so no replicate is discarded. The cluster unit is stated in each table: tower for the tower and tower14 splits and for the original-T3 audit, and \texttt{parent\_id} for the image split. Because Holm families differ across tables, the same contrast can carry different adjusted values in different tables; the family is stated in every caption.

All experiments ran on one NVIDIA RTX PRO 6000 Blackwell (96 GB) and consumed approximately 205 GPU-hours. Table A5 gives the software and run configuration and Table A6 the experiment ledger.

\section*{5. Results}
Balanced accuracies are reported to three decimal places, or as percentages to two decimal places. The sampling uncertainty of a balanced accuracy at n = 2,021 is of order ±1 percentage point, so differences below roughly two points are not interpreted without an accompanying paired test.

\subsection*{5.1. Zero-Shot Error Profiles}
Zero-shot T2 balanced accuracy spans 42.0\% to 74.3\% across seven models. Format-failure rate was 0.0\% throughout, which excludes parsing failure as an explanation for any low score below. It does not exclude refusal, which is a legal and well-formed output in this benchmark and is scored separately.

\begin{table}[htbp]
\centering
\setlength{\tabcolsep}{3.0pt}
\renewcommand{\arraystretch}{1.15}
\caption*{Table 2. Zero-shot T2 performance (n = 2,021; 1,255 normal, 766 defective). Abstention rates are measured from stored predictions for the five locally run models and derived from Equation (1) for the two API-served models.}
\begin{tabularx}{\textwidth}{>{\hsize=1.402\hsize\RaggedRight\hyphenpenalty=10000\exhyphenpenalty=50\arraybackslash}X >{\hsize=1.033\hsize\RaggedRight\hyphenpenalty=10000\exhyphenpenalty=50\arraybackslash}X >{\hsize=0.642\hsize\RaggedRight\hyphenpenalty=10000\exhyphenpenalty=50\arraybackslash}X >{\hsize=0.741\hsize\RaggedRight\hyphenpenalty=10000\exhyphenpenalty=50\arraybackslash}X >{\hsize=1.484\hsize\RaggedRight\hyphenpenalty=10000\exhyphenpenalty=50\arraybackslash}X >{\hsize=0.849\hsize\RaggedRight\hyphenpenalty=10000\exhyphenpenalty=50\arraybackslash}X >{\hsize=0.849\hsize\RaggedRight\hyphenpenalty=10000\exhyphenpenalty=50\arraybackslash}X}
\toprule
Model & Parameters & BalAcc & β (miss) & α (hallucination) & γ\textsubscript{D} & γ\textsubscript{N} \\
\midrule
InternVL3.5-2B & 2B & 0.420 & 0.00\% ‡ & 70.12\% & 20.10\% & 25.82\% \\
Qwen3-VL-2B & 2B & 0.447 & 1.04\% & 60.32\% & 22.19\% & 27.09\% \\
Qwen2.5-VL-7B & 7B & 0.578 & 8.88\% & 31.79\% & 20.89\% & 22.79\% \\
Qwen3.8-27B & 27B (API) & 0.665 & 64.49\% & 0.96\% & ≤1.55\% † & ≤1.55\% † \\
InternVL3.5-8B & 8B & 0.666 & 63.32\% & 0.56\% & 1.83\% & 1.20\% \\
Qwen3.5-122B & 122B (API) & 0.710 & 45.04\% & 4.14\% & ≤8.76\% † & ≤8.76\% † \\
Qwen3-VL-8B & 8B & 0.743 & 39.82\% & 2.07\% & 4.83\% & 4.78\% \\
\bottomrule
\end{tabularx}
\end{table}

‡ Zero misses observed among 766 defective items; one-sided 95\% upper bound 0.39\%. † Derived bound: the API result records do not store per-class predictions, so Equation (1) identifies only the sum γ\textsubscript{D} + γ\textsubscript{N}.

Miss and hallucination rates invert with scale. The 2B models approximate an always-defect policy, with α between 60\% and 70\% and β near zero; the 8B to 27B models show the reverse, with α below 5\% and β between 40\% and 65\%; and the 7B model lies between them. Figure 1 plots the (α, β) plane directly.

\begin{figure}[!htbp]
\centering
\IfFileExists{figures/figure1.jpg}{\includegraphics[width=0.95\linewidth]{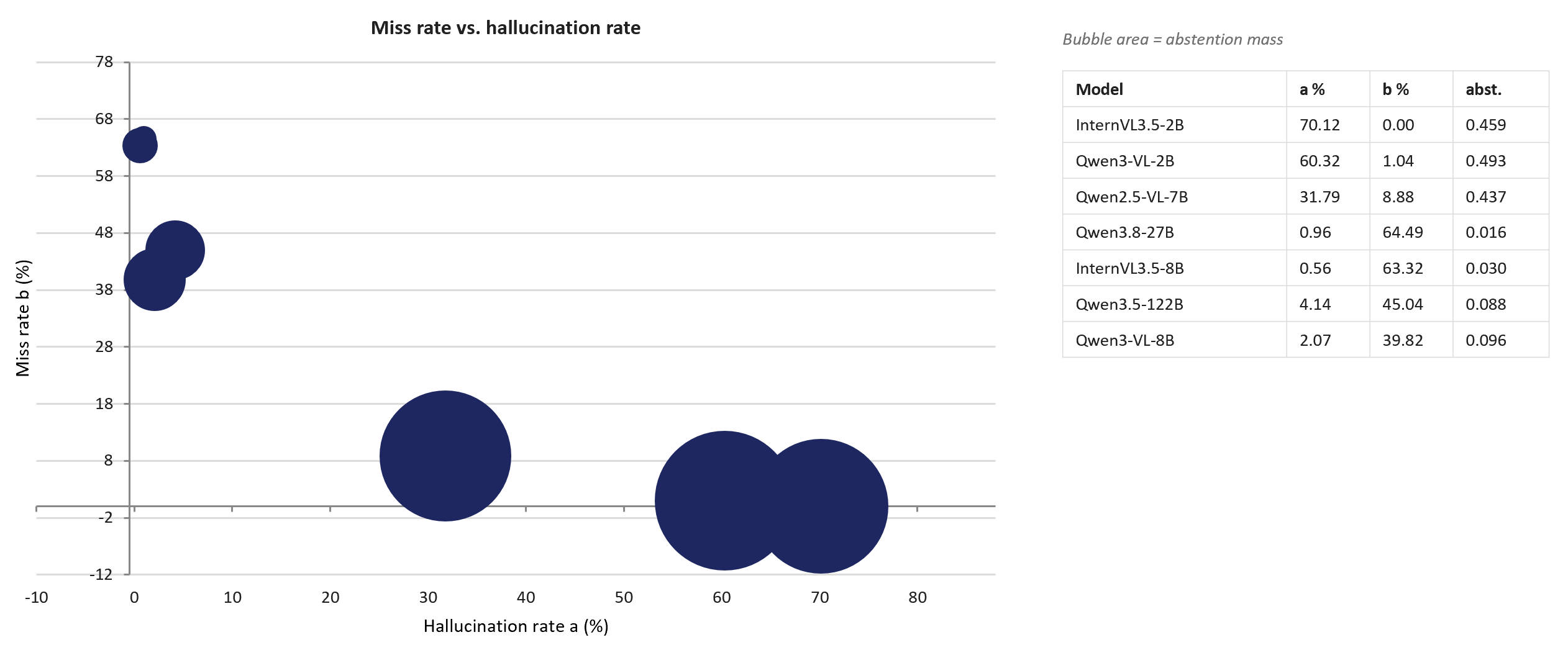}}{\fbox{\parbox{0.92\linewidth}{\centering\vspace{2.5cm}\small Placeholder for \texttt{figures/figure1.jpg}\vspace{2.5cm}}}}
\caption*{Figure 1. Zero-shot error-regime plane. Hallucination rate α against miss rate β for the seven zero-shot models of Table 2, with marker area encoding total abstention. The plot separates the miss-dominated, hallucination-dominated, and abstention-dominated regimes that balanced accuracy conflates.}
\end{figure}
\FloatBarrier

The abstention columns provide an independent check on Equation (1). For every locally run model the measured γ\textsubscript{D} + γ\textsubscript{N} reproduces the value implied by BalAcc, α, and β to within rounding: 0.4592 for InternVL3.5-2B, 0.4928 for Qwen3-VL-2B, 0.4368 for Qwen2.5-VL-7B, 0.0303 for InternVL3.5-8B, and 0.0961 for Qwen3-VL-8B.

The consequence for interpretation is direct. InternVL3.5-2B, whose headline property is a zero miss rate, leaves 20.1\% of defective items unjudged. Its β ≈ 0 does not mean that it never misses a defect, but that among the defective items on which it committed to an answer it never answered normal, which an inspection workflow would price very differently. The three lowest-scoring models in Table 2 are also the three with the largest abstention mass, so the abstention-dominated case of Proposition 1 is the most common behaviour of the small models rather than a theoretical curiosity. By the same identity, the three adapted models of Table 4 have γ\textsubscript{D} + γ\textsubscript{N} = 0.000 to within rounding: adaptation removes abstention entirely, which is expected because no training item carries it as a target.

\subsection*{5.2. Cost-Sensitive Ranking and Its Assumptions}
Applying Algorithm 2 to InternVL3.5-2B against InternVL3.5-8B yields the crossover surface of Table 3.

\begin{table}[htbp]
\centering
\setlength{\tabcolsep}{3.0pt}
\renewcommand{\arraystretch}{1.15}
\caption*{Table 3. Crossover cost ratio ρ* over defect prevalence π\textsubscript{D} and abstention cost κ, computed from the measured abstention rates of Table 2.}
\begin{tabularx}{0.769\textwidth}{>{\hsize=2.809\hsize\RaggedRight\hyphenpenalty=10000\exhyphenpenalty=50\arraybackslash}X >{\hsize=0.460\hsize\RaggedRight\hyphenpenalty=10000\exhyphenpenalty=50\arraybackslash}X >{\hsize=0.678\hsize\RaggedRight\hyphenpenalty=10000\exhyphenpenalty=50\arraybackslash}X >{\hsize=0.593\hsize\RaggedRight\hyphenpenalty=10000\exhyphenpenalty=50\arraybackslash}X >{\hsize=0.460\hsize\RaggedRight\hyphenpenalty=10000\exhyphenpenalty=50\arraybackslash}X}
\toprule
π\textsubscript{D} & κ = 0 & κ = 0.25 & κ = 0.5 & κ = 1 \\
\midrule
0.379 (this benchmark) & 1.8 & 2.0 & 2.3 & 2.7 \\
0.10 & 9.9 & 10.8 & 11.8 & 13.7 \\
0.05 & 20.9 & 22.8 & 24.7 & 28.5 \\
0.01 (low-prevalence sensitivity point) & 108.8 & 118.5 & 128.1 & 147.5 \\
\bottomrule
\end{tabularx}
\end{table}

Two readings follow. Pricing manual review at parity with a false alarm raises the crossover by 50\%, from 1.8 to 2.7, but does not abolish it, because the near-zero miss rate of the 2B model remains an advantage once the review cost is paid. Prevalence has the larger effect. At a 1\% defect rate the 42\%-accuracy model becomes preferable only if a missed defect costs more than roughly a hundred false alarms, which is a sensitivity result rather than a statement about routine operating assumptions. The defensible scope of the finding is that at this benchmark's constructed 37.9\% prevalence the accuracy ranking and the risk ranking invert at a cost ratio between 1.8 and 2.7 depending on how review is priced, and that at plausible field prevalence they do not invert at any cost ratio we would defend as routine. Figure 2 shows the surface.

\begin{figure}[!htbp]
\centering
\IfFileExists{figures/figure2.jpg}{\includegraphics[width=0.95\linewidth]{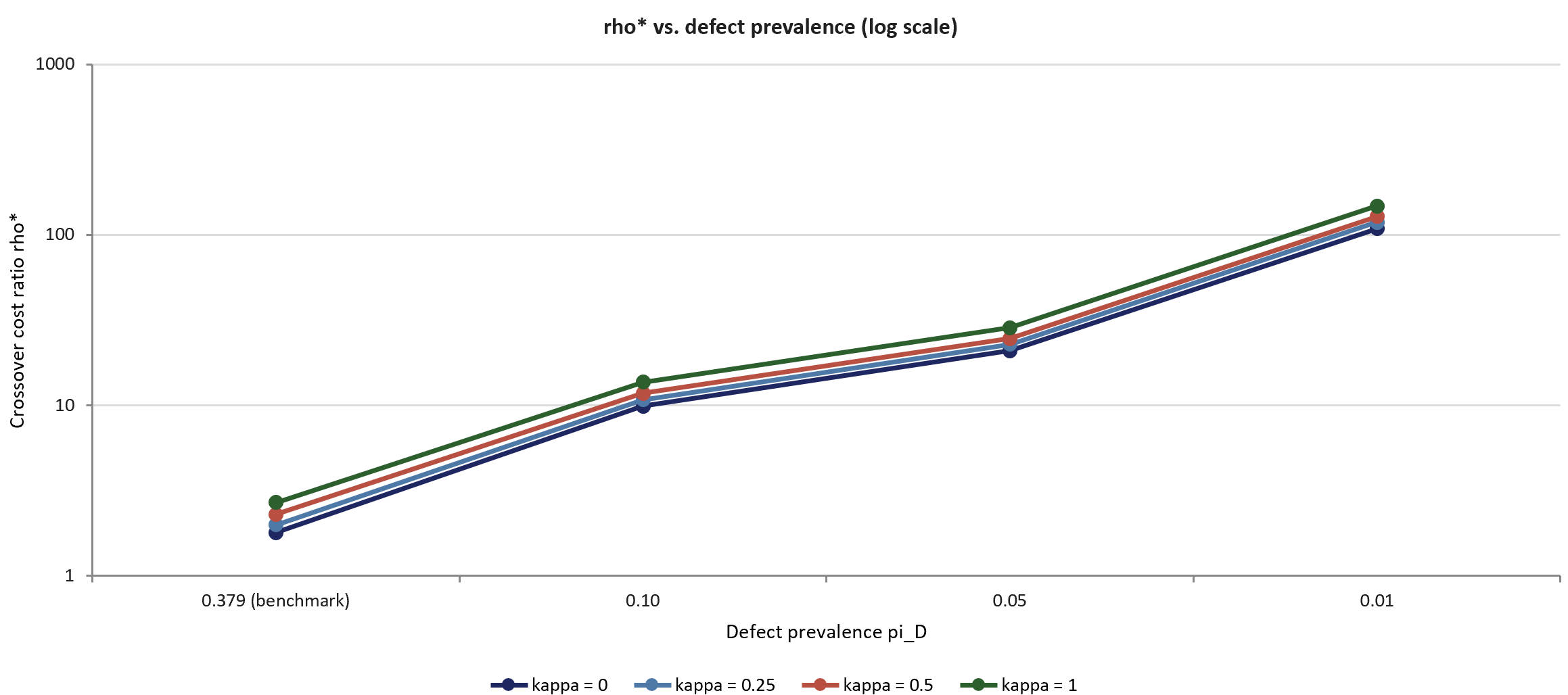}}{\fbox{\parbox{0.92\linewidth}{\centering\vspace{2.5cm}\small Placeholder for \texttt{figures/figure2.jpg}\vspace{2.5cm}}}}
\caption*{Figure 2. Crossover cost ratio ρ* as a function of defect prevalence π\textsubscript{D} and abstention-cost ratio κ, from Table 3, on a linear κ axis because κ = 0 is included. The surface shows how prevalence and manual-review cost jointly move the crossover.}
\end{figure}
\FloatBarrier

The error-asymmetry ratio EAR = (β + ε)/(α + ε) is a compact ordering device for the (α, β) plane but is unstable in the smoothing constant ε, which is a modelling choice rather than a measurement. Across ε ∈ \{0.1, 0.01, 0.001\} the value for InternVL3.5-8B varies by a factor of fourteen, and the ordering of the 27B and 122B models is not preserved (Table A7). EAR also ignores abstention entirely. It is therefore reported only alongside the abstention columns, and no regression is fitted through it.

\subsection*{5.3. Adaptation}
\begin{table}[htbp]
\centering
\setlength{\tabcolsep}{4pt}
\renewcommand{\arraystretch}{1.15}
\caption*{Table 4. LoRA-adapted models on the full T2 (n = 2,021) and T3 (n = 7,284) test sets, seed 42.}
\begin{tabularx}{\textwidth}{>{\hsize=1.631\hsize\RaggedRight\hyphenpenalty=10000\exhyphenpenalty=50\arraybackslash}X >{\hsize=0.895\hsize\RaggedRight\hyphenpenalty=10000\exhyphenpenalty=50\arraybackslash}X >{\hsize=0.524\hsize\RaggedRight\hyphenpenalty=10000\exhyphenpenalty=50\arraybackslash}X >{\hsize=0.524\hsize\RaggedRight\hyphenpenalty=10000\exhyphenpenalty=50\arraybackslash}X >{\hsize=1.162\hsize\RaggedRight\hyphenpenalty=10000\exhyphenpenalty=50\arraybackslash}X >{\hsize=1.172\hsize\RaggedRight\hyphenpenalty=10000\exhyphenpenalty=50\arraybackslash}X >{\hsize=1.092\hsize\RaggedRight\hyphenpenalty=10000\exhyphenpenalty=50\arraybackslash}X}
\toprule
Model & T2 BalAcc & T2 β & T2 α & T2 abstention & T3 BalAcc (K = 7) & T3 macro-F1 \\
\midrule
InternVL3.5-2B + LoRA & 0.949 & 9.79\% & 0.40\% & 0.000 & 0.742 & 0.772 \\
InternVL3.5-8B + LoRA & 0.959 & 7.57\% & 0.72\% & 0.000 & 0.779 & 0.795 \\
Qwen3-VL-8B + LoRA & 0.971 & 5.09\% & 0.64\% & 0.000 & 0.891 & 0.919 \\
\bottomrule
\end{tabularx}
\end{table}

The seven-way scores are obtained by mapping each model's answer-option output back to the seven ground-truth labels. Because the three-way scores of Section 5.6 average over a different partition rather than a collapse of this one, they are not constrained to exceed these values, and the three-way score of 0.881 for Qwen3-VL-8B against a seven-way score of 0.891 is not an inconsistency. Macro recall and macro-F1 can rank models differently because the first is recall-only whereas the second also includes precision, so a model with higher per-class recall but more cross-class false positives can have lower macro-F1.

\begin{table}[htbp]
\centering
\setlength{\tabcolsep}{3.0pt}
\renewcommand{\arraystretch}{1.15}
\caption*{Table 5. Global two-seed replication of the two 8B adapted VLMs. The T3 column is seven-class macro recall obtained by mapping option outputs back to labels.}
\begin{tabularx}{0.870\textwidth}{>{\hsize=1.282\hsize\RaggedRight\hyphenpenalty=10000\exhyphenpenalty=50\arraybackslash}X >{\hsize=0.614\hsize\RaggedRight\hyphenpenalty=10000\exhyphenpenalty=50\arraybackslash}X >{\hsize=0.835\hsize\RaggedRight\hyphenpenalty=10000\exhyphenpenalty=50\arraybackslash}X >{\hsize=1.269\hsize\RaggedRight\hyphenpenalty=10000\exhyphenpenalty=50\arraybackslash}X}
\toprule
Model & Seed & T2 BalAcc & T3 macro recall (K = 7) \\
\midrule
Qwen3-VL-8B + LoRA & 42 & 0.9714 & 0.8908 \\
Qwen3-VL-8B + LoRA & 41 & 0.9800 & 0.9010 \\
Qwen3-VL-8B + LoRA & mean ± SD & 0.9757 ± 0.0061 & 0.8959 ± 0.0072 \\
InternVL3.5-8B + LoRA & 42 & 0.9586 & 0.7790 \\
InternVL3.5-8B + LoRA & 41 & 0.9586 & 0.7440 \\
\bottomrule
\end{tabularx}
\end{table}

The seed-41 Qwen evaluation has no format failures on either task. Its option-level T3 balanced accuracy is 0.9379, and after mapping options to the seven labels its macro recall is 0.9010. The replication changes Qwen T2 by 0.86 points and seven-way macro recall by 1.02 points, so the global Qwen result is not a single-seed point estimate. Two seeds remain too few for a precise seed-variance estimate, and the InternVL seven-way seed difference of 3.50 points is more than three times the Qwen one, which is itself a caution against reading either model's single-seed seven-way value as stable. Figure 3 shows both models across seeds.

\begin{figure}[!htbp]
\centering
\IfFileExists{figures/figure3.jpg}{\includegraphics[width=0.95\linewidth]{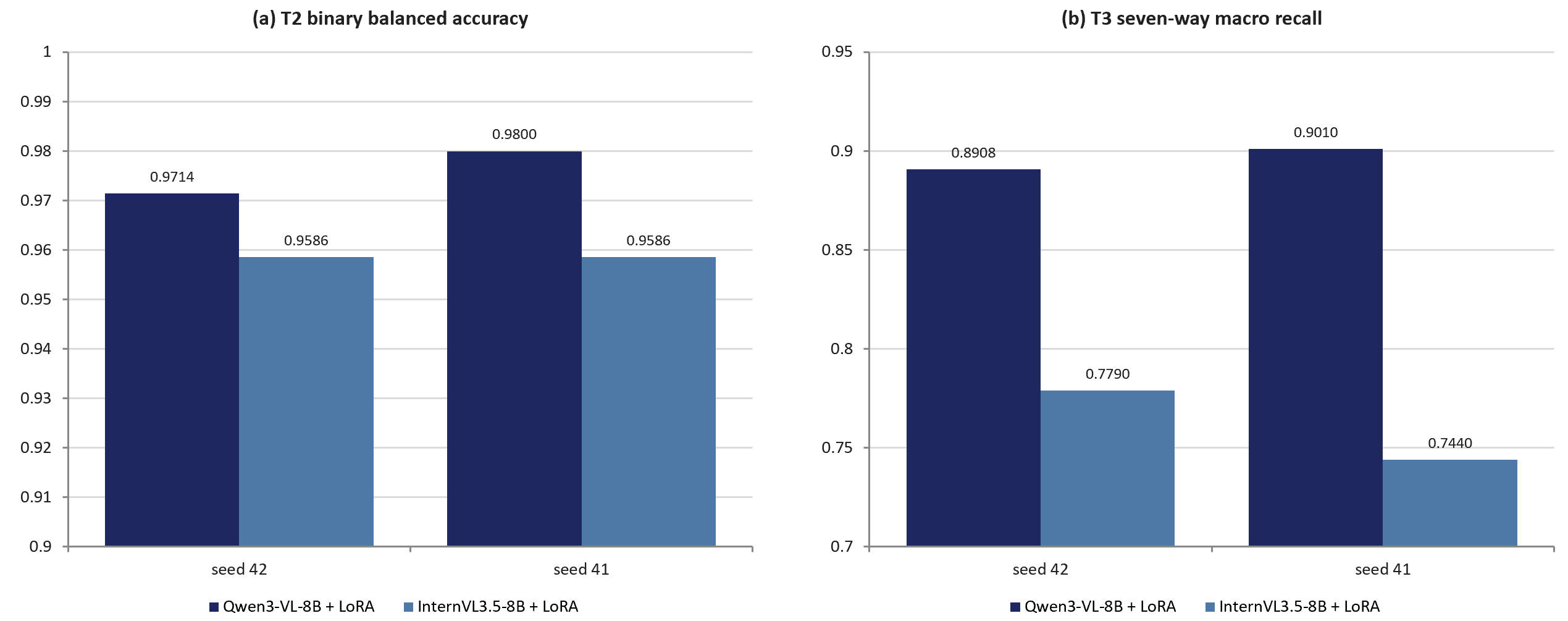}}{\fbox{\parbox{0.92\linewidth}{\centering\vspace{2.5cm}\small Placeholder for \texttt{figures/figure3.jpg}\vspace{2.5cm}}}}
\caption*{Figure 3. Global two-seed replication. Binary balanced accuracy and seven-way macro recall for Qwen3-VL-8B and InternVL3.5-8B at seeds 41 and 42, from Table 5, in two panels.}
\end{figure}
\FloatBarrier

Two features of Table 4 motivate Section 5.7. First, the binary and seven-way orderings differ: at K = 2 the three models span 2.2 points, while at K = 7 they span 14.9 points. Second, Qwen3-VL-8B is 11.18 points above InternVL3.5-8B at seven-way typing but only 1.3 points above it at binary screening.

\begin{table}[htbp]
\centering
\setlength{\tabcolsep}{4pt}
\renewcommand{\arraystretch}{1.15}
\caption*{Table 6. McNemar's exact test on paired T2 predictions (n = 2,021). Holm family: the three comparisons in this table.}
\begin{tabularx}{\textwidth}{>{\hsize=1.312\hsize\RaggedRight\hyphenpenalty=10000\exhyphenpenalty=50\arraybackslash}X >{\hsize=1.222\hsize\RaggedRight\hyphenpenalty=10000\exhyphenpenalty=50\arraybackslash}X >{\hsize=0.955\hsize\RaggedRight\hyphenpenalty=10000\exhyphenpenalty=50\arraybackslash}X >{\hsize=0.919\hsize\RaggedRight\hyphenpenalty=10000\exhyphenpenalty=50\arraybackslash}X >{\hsize=0.770\hsize\RaggedRight\hyphenpenalty=10000\exhyphenpenalty=50\arraybackslash}X >{\hsize=0.780\hsize\RaggedRight\hyphenpenalty=10000\exhyphenpenalty=50\arraybackslash}X >{\hsize=1.042\hsize\RaggedRight\hyphenpenalty=10000\exhyphenpenalty=50\arraybackslash}X}
\toprule
Comparison & Discordant pairs (b, c) & Δ accuracy & 95\% CI & p (exact) & p (Holm) & Verdict \\
\midrule
InternVL-2B vs. Qwen3-VL-8B & 21, 54 & −1.63 pt & [−2.47, −0.80] & 0.0002 & 0.0005 & significant \\
InternVL-8B vs. Qwen3-VL-8B & 20, 40 & −0.99 pt & [−1.74, −0.24] & 0.0135 & 0.0270 & significant \\
InternVL-8B vs. InternVL-2B & 26, 13 & +0.64 pt & [−0.01, +1.29] & 0.0533 & 0.0533 & not detected \\
\bottomrule
\end{tabularx}
\end{table}

The Holm correction changes no verdict. The 2B-versus-8B comparison does not reach significance, and its confidence interval, which includes zero but is bounded above at 1.3 points, is the substantive result: the experiment does not distinguish a difference smaller than about 1.3 percentage points from noise. Relative to zero-shot, adaptation moves InternVL3.5-2B from 42.0\% to 94.9\%, InternVL3.5-8B from 66.6\% to 95.9\%, and Qwen3-VL-8B from 74.3\% to 97.1\%. The miss rate falls from the 40\% to 65\% range to the 5\% to 10\% range, hallucination stays below 1\%, and abstention falls to zero.

\subsection*{5.4. Multi-Task Mixing and Epoch Budget}
\begin{table}[htbp]
\centering
\setlength{\tabcolsep}{3.0pt}
\renewcommand{\arraystretch}{1.15}
\caption*{Table 7. Training corpus and epoch budget, InternVL3.5-2B + LoRA, evaluated on the shared 200-item subsample S200. The T3 diagnostic for these configurations covers only five of the seven classes and is reported separately in Table A8.}
\begin{tabularx}{0.732\textwidth}{>{\hsize=1.378\hsize\RaggedRight\hyphenpenalty=10000\exhyphenpenalty=50\arraybackslash}X >{\hsize=1.158\hsize\RaggedRight\hyphenpenalty=10000\exhyphenpenalty=50\arraybackslash}X >{\hsize=0.750\hsize\RaggedRight\hyphenpenalty=10000\exhyphenpenalty=50\arraybackslash}X >{\hsize=1.073\hsize\RaggedRight\hyphenpenalty=10000\exhyphenpenalty=50\arraybackslash}X >{\hsize=0.750\hsize\RaggedRight\hyphenpenalty=10000\exhyphenpenalty=50\arraybackslash}X >{\hsize=0.891\hsize\RaggedRight\hyphenpenalty=10000\exhyphenpenalty=50\arraybackslash}X}
\toprule
Configuration & Train items & Epochs & T2 BalAcc & T2 β & Eval loss \\
\midrule
T2-only & 6,949 & 3 & 0.908 & 17.57\% & — \\
T3-only & 25,460 & 3 & 0.874 & 24.32\% & — \\
Mixed & 32,409 & 3 & 0.932 & 13.51\% & 0.035 \\
Mixed & 32,409 & 10 & 0.935 & 12.16\% & 0.099 \\
\bottomrule
\end{tabularx}
\end{table}

The mixed corpus outperforms both specialists on T2. The margin over the T2-only specialist is 2.4 points and over the T3-only specialist 5.8 points; the exact per-item counts are available in the released prediction files. The result is consistent with cross-task regularization \cite{ref27,ref28}, although a training-volume contribution cannot be separated without a volume-matched control.

Extending training from three to ten epochs leaves T2 essentially unchanged while evaluation loss rises from 0.035 to 0.099, and the five-class T3 diagnostic falls by 10.1 points (Table A8). The pattern is consistent with over-fitting confined to the task with more classes and thinner per-class support, and the loss signal is independent of the subsample. We therefore use three epochs throughout, and recommend that early stopping in multi-task inspection pipelines be governed by the hardest task in the mixture rather than the headline one. Separately, at four examples per class, raising the epoch budget from 10 to 100 changes T2 accuracy by at most 1.0 point non-monotonically while wall-clock cost rises 8.6-fold, so additional optimization steps do not substitute for additional examples.

Direct measurement of S200 gives 126 normal and 74 defective items at T2, and 177 \texttt{good}, 12 \texttt{rust}, 7 \texttt{missing-cap}, 3 \texttt{corrosão}, and 1 \texttt{nest} item at T3, with \texttt{peeling-paint} and \texttt{torned-up} absent. The S200 T3 results are therefore a five-class diagnostic and are not comparable to any seven-class value elsewhere in this paper.

\subsection*{5.5. Partition Parity at Binary Screening}
An initial comparison scored the vision backbones on a superseded 7,851-item seven-class file and the VLMs on the 2,021-item binary T2 file, failing Algorithm 3 at lines 2 and 3, and produced a VLM lead of 13 to 15 points. Attribution analysis shows that the two files have identical label and asset distributions with no crop-identifier overlap, so the 28-point difference in the vision backbone's score between them is attributable to training-configuration differences rather than to the partitions themselves; the earlier file predates the pipeline of Section 4.6. That comparison is retained only as the object on which the parity check fails.

\begin{table}[htbp]
\centering
\setlength{\tabcolsep}{3.0pt}
\renewcommand{\arraystretch}{1.15}
\caption*{Table 8. Vision-only backbones across training budget on the T2 test partition (n = 2,021), three seeds, mean ± SD.}
\begin{tabularx}{0.514\textwidth}{>{\hsize=1.284\hsize\RaggedRight\hyphenpenalty=10000\exhyphenpenalty=50\arraybackslash}X >{\hsize=0.858\hsize\RaggedRight\hyphenpenalty=10000\exhyphenpenalty=50\arraybackslash}X >{\hsize=0.858\hsize\RaggedRight\hyphenpenalty=10000\exhyphenpenalty=50\arraybackslash}X}
\toprule
Examples per class & ResNet-50 & Swin-T \\
\midrule
4 & 61.48 ± 4.15 & 71.66 ± 8.92 \\
8 & 67.14 ± 0.29 & 72.65 ± 8.30 \\
16 & 72.70 ± 1.07 & 81.45 ± 6.36 \\
32 & 81.59 ± 0.67 & 89.30 ± 1.51 \\
64 & 89.56 ± 1.13 & 90.80 ± 1.25 \\
128 & 92.24 ± 0.26 & 93.88 ± 0.73 \\
all & 96.54 ± 0.10 & 97.17 ± 0.23 \\
\bottomrule
\end{tabularx}
\end{table}

Swin-T leads ResNet-50 at every budget, with the margin largest in the few-shot regime and closing to 0.6 points at full data. A cross-family few-shot comparison is not reported, because the available VLM ladder was scored on a 200-item subsample against the backbones' full partition with unequal replication, failing Algorithm 3 at lines 4 and 5.

\begin{table}[htbp]
\centering
\setlength{\tabcolsep}{4pt}
\renewcommand{\arraystretch}{1.15}
\caption*{Table 9. Cross-family comparison at full training data on the identical T2 test file (n = 2,021), with paired McNemar tests against the strongest adapted VLM. Algorithm 3: PASS at lines 1 to 4, PASS with caveat at lines 5 and 6.}
\begin{tabularx}{\textwidth}{>{\hsize=1.342\hsize\RaggedRight\hyphenpenalty=10000\exhyphenpenalty=50\arraybackslash}X >{\hsize=1.500\hsize\RaggedRight\hyphenpenalty=10000\exhyphenpenalty=50\arraybackslash}X >{\hsize=0.885\hsize\RaggedRight\hyphenpenalty=10000\exhyphenpenalty=50\arraybackslash}X >{\hsize=0.413\hsize\RaggedRight\hyphenpenalty=10000\exhyphenpenalty=50\arraybackslash}X >{\hsize=1.213\hsize\RaggedRight\hyphenpenalty=10000\exhyphenpenalty=50\arraybackslash}X >{\hsize=0.647\hsize\RaggedRight\hyphenpenalty=10000\exhyphenpenalty=50\arraybackslash}X}
\toprule
Model & Family & BalAcc & Seeds & Discordant pairs vs. Qwen3-VL-8B & p (exact) \\
\midrule
Swin-T & vision-only (transformer) & 97.17 ± 0.23 & 3 & 24, 24 & 1.000 \\
Qwen3-VL-8B + LoRA & VLM & 97.14 & 1 & — & — \\
ResNet-50 & vision-only (convolutional) & 96.54 ± 0.10 & 3 & 22, 30 & 0.332 \\
InternVL3.5-8B + LoRA & VLM & 95.86 & 1 & — & — \\
InternVL3.5-2B + LoRA & VLM & 94.91 & 1 & — & — \\
\bottomrule
\end{tabularx}
\end{table}

At full training data on a matched partition, the highest-scoring vision-only configuration and the best adapted VLM are separated by 0.03 percentage points, and McNemar's test returns p = 1.000: the two models disagree on 48 items and split them 24 to 24. This is a balanced discordant-pair result rather than a formal equivalence test, and no practical equivalence margin was pre-specified. The finding that survives the audit is the removal of the large lead visible in the uncontrolled comparison, not evidence that the two families are equivalent.

\subsection*{5.6. Label-Space Parity: Seven-Way Defect Typing}
\begin{table}[htbp]
\centering
\setlength{\tabcolsep}{4pt}
\renewcommand{\arraystretch}{1.15}
\caption*{Table 10. Balanced accuracy (\%) across label-space granularity. All rows are scored on the stated file with the stated ground-truth class set.}
\begin{tabularx}{\textwidth}{>{\hsize=0.178\hsize\RaggedRight\hyphenpenalty=10000\exhyphenpenalty=50\arraybackslash}X >{\hsize=0.894\hsize\RaggedRight\hyphenpenalty=10000\exhyphenpenalty=50\arraybackslash}X >{\hsize=1.192\hsize\RaggedRight\hyphenpenalty=10000\exhyphenpenalty=50\arraybackslash}X >{\hsize=0.998\hsize\RaggedRight\hyphenpenalty=10000\exhyphenpenalty=50\arraybackslash}X >{\hsize=1.290\hsize\RaggedRight\hyphenpenalty=10000\exhyphenpenalty=50\arraybackslash}X >{\hsize=1.290\hsize\RaggedRight\hyphenpenalty=10000\exhyphenpenalty=50\arraybackslash}X >{\hsize=1.158\hsize\RaggedRight\hyphenpenalty=10000\exhyphenpenalty=50\arraybackslash}X}
\toprule
K & Test file, n & ResNet-50 (224 px) & Swin-T (224 px) & InternVL3.5-2B & InternVL3.5-8B & Qwen3-VL-8B \\
\midrule
2 & T2, 2,021 & 96.54 & 97.17 & 94.91 & 95.86 & 97.14 \\
3 & T3, 7,284 & 92.26 & 93.09 & 85.77 & 89.17 & 88.10 \\
7 & T3, 7,284 & 57.37 & 54.23 & 74.16 & 77.90 & 89.08 \\
\bottomrule
\end{tabularx}
\end{table}

At the conventional 224 px input the seven-way row reverses the sign of the binary row. Against InternVL3.5-8B the 224 px gaps are 20.53 points for ResNet-50 and 23.67 points for Swin-T; against Qwen3-VL-8B they are 31.71 and 34.85 points. Section 5.7 shows that pixel budget accounts for most of the InternVL gap but not for the difference between the two VLMs. The three-way row is reported for completeness: as noted in Section 4.6 it averages over the option-level partition rather than a collapse of the seven-way label space, so it is comparable within itself but not to the row below it.

Inspection of the vision-only confusion matrices at 224 px shows the majority of \texttt{rust} instances predicted as \texttt{good}. The two classes are similar in raw pixel statistics, surface discoloration against normal patina and lighting variation, and at 224 px a discriminative backbone has neither the spatial detail nor the definitional prior to separate them. Both deficits are addressed experimentally below.

\subsection*{5.7. Separating Pixel Budget, Language Grounding, Scale, and Model Identity}
\subsubsection*{5.7.1. Resolution and Pixel-Budget Control}
\begin{table}[htbp]
\centering
\setlength{\tabcolsep}{3.0pt}
\renewcommand{\arraystretch}{1.15}
\caption*{Table 11. Vision-only balanced accuracy (\%) at K = 7 across input resolution (T3 test, n = 7,284), with paired McNemar tests on raw per-item correctness against InternVL3.5-8B at native tiling (77.90\%).}
\begin{tabularx}{\textwidth}{>{\hsize=1.426\hsize\RaggedRight\hyphenpenalty=10000\exhyphenpenalty=50\arraybackslash}X >{\hsize=0.419\hsize\RaggedRight\hyphenpenalty=10000\exhyphenpenalty=50\arraybackslash}X >{\hsize=1.287\hsize\RaggedRight\hyphenpenalty=10000\exhyphenpenalty=50\arraybackslash}X >{\hsize=0.529\hsize\RaggedRight\hyphenpenalty=10000\exhyphenpenalty=50\arraybackslash}X >{\hsize=1.339\hsize\RaggedRight\hyphenpenalty=10000\exhyphenpenalty=50\arraybackslash}X}
\toprule
Vision-only configuration & BalAcc & Discordant pairs (b, c) & p (exact) & Verdict on raw accuracy \\
\midrule
ResNet-50, 224 px & 57.37 & — & — & — \\
ResNet-50, 448 px & 81.08 & 75, 37 & 0.0004 & vision-only ahead \\
ResNet-50, 672 px & 78.47 & 70, 54 & 0.178 & not detected \\
ResNet-50, 896 px & 76.10 & 74, 98 & 0.079 & not detected \\
Swin-T, 224 px & 54.23 & — & — & — \\
Swin-T, 448 px & 75.11 & 66, 29 & 0.0002 & vision-only ahead \\
Swin-T, 672 px & 73.20 & 67, 53 & 0.235 & not detected \\
\bottomrule
\end{tabularx}
\end{table}

Raising ResNet-50 from 224 px to 448 px adds 23.71 points at K = 7, and raising Swin-T adds 20.88 points. The comparison of interest is the one at the measured matched budget. Section 4.6 records a mean of 2.94 patches per T3 item for InternVL3.5-8B at native settings, an aggregate budget equivalent to a 768 px square image, which lies between the 672 px and 896 px settings. At those two settings ResNet-50 scores 78.47\% and 76.10\% and Swin-T scores 73.20\%, against 77.34\% to 77.90\% for InternVL3.5-8B across evaluation passes, so InternVL falls between the two ResNet-50 values and above the Swin-T value. Expressed within a single backbone, the gap against InternVL3.5-8B changes from +20.53 to −0.57 points for ResNet-50 and from +23.67 to +4.70 points for Swin-T when moving from 224 px to 672 px.

The 448 px results are the highest-scoring point of a four-point sweep evaluated on the test set, so that setting carries a selection advantage which the VLMs, evaluated only in fixed configurations, do not. The 672 px and 896 px rows are the controlled comparison.

A paired significance test on the Qwen comparison is reported in Table 12. It illustrates the divergence between raw accuracy and macro recall rather than establishing the gap.

\begin{table}[htbp]
\centering
\setlength{\tabcolsep}{3.0pt}
\renewcommand{\arraystretch}{1.15}
\caption*{Table 12. McNemar's exact test on raw per-item correctness, Qwen3-VL-8B against vision-only configurations, T3 test (n = 7,284).}
\begin{tabularx}{\textwidth}{>{\hsize=1.687\hsize\RaggedRight\hyphenpenalty=10000\exhyphenpenalty=50\arraybackslash}X >{\hsize=0.772\hsize\RaggedRight\hyphenpenalty=10000\exhyphenpenalty=50\arraybackslash}X >{\hsize=0.428\hsize\RaggedRight\hyphenpenalty=10000\exhyphenpenalty=50\arraybackslash}X >{\hsize=1.113\hsize\RaggedRight\hyphenpenalty=10000\exhyphenpenalty=50\arraybackslash}X}
\toprule
Comparison & Macro-recall gap & p (exact) & Verdict on raw accuracy \\
\midrule
Qwen3-VL-8B vs. ResNet-50, 448 px & 8.0 pt & 0.059 & not detected \\
Qwen3-VL-8B vs. ResNet-50, 672 px & 10.6 pt & 0.923 & not detected \\
Qwen3-VL-8B vs. ResNet-50, 896 px & 12.9 pt & 0.002 & detected \\
Qwen3-VL-8B vs. Swin-T, 448 px & 14.0 pt & 0.078 & not detected \\
\bottomrule
\end{tabularx}
\end{table}

The pattern is not monotonic in the macro-recall gap. The 672 px row has a large macro-recall gap and the weakest raw-accuracy signal, with p = 0.923. Because \texttt{good} constitutes 89.5\% of the T3 test set, raw per-item correctness is dominated by a class on which both models score near ceiling, and a macro-recall difference concentrated in rare classes can be nearly invisible to a test built on item-level pairing. A tower-clustered macro-recall test therefore replaces these raw-accuracy tests as the inferential summary.

\begin{table}[htbp]
\centering
\setlength{\tabcolsep}{3.0pt}
\renewcommand{\arraystretch}{1.15}
\caption*{Table 13. Tower-clustered paired macro-recall tests on the original T3 comparison (10,000 sign-flip replicates). Holm family: the five vision-only configurations in this table.}
\begin{tabularx}{\textwidth}{>{\hsize=1.868\hsize\RaggedRight\hyphenpenalty=10000\exhyphenpenalty=50\arraybackslash}X >{\hsize=1.167\hsize\RaggedRight\hyphenpenalty=10000\exhyphenpenalty=50\arraybackslash}X >{\hsize=1.084\hsize\RaggedRight\hyphenpenalty=10000\exhyphenpenalty=50\arraybackslash}X >{\hsize=0.469\hsize\RaggedRight\hyphenpenalty=10000\exhyphenpenalty=50\arraybackslash}X >{\hsize=0.535\hsize\RaggedRight\hyphenpenalty=10000\exhyphenpenalty=50\arraybackslash}X >{\hsize=0.877\hsize\RaggedRight\hyphenpenalty=10000\exhyphenpenalty=50\arraybackslash}X}
\toprule
Comparison & Macro-recall gap & 95\% CI & Raw p & Holm p & Verdict \\
\midrule
Qwen vs. ResNet-50, 448 px & +7.99 pt & [−4.59, +15.01] & 0.287 & 0.834 & not detected \\
Qwen vs. ResNet-50, 672 px & +10.61 pt & [−5.28, +28.11] & 0.352 & 0.834 & not detected \\
Qwen vs. ResNet-50, 896 px & +12.93 pt & [−3.57, +30.57] & 0.278 & 0.834 & not detected \\
Qwen vs. Swin-T, 448 px & +13.95 pt & [+0.31, +28.24] & 0.101 & 0.405 & not detected \\
Qwen vs. Swin-T, 672 px & +15.90 pt & [+2.17, +29.71] & 0.0079 & 0.0395 & detected \\
\bottomrule
\end{tabularx}
\end{table}

The Qwen–ResNet-50 448 px effect is +7.99 points but is not statistically established after tower clustering. The contrast against Swin-T at 672 px is the only one in this family that survives Holm correction. Table 14 gives the corresponding per-class breakdown.

\begin{table}[htbp]
\centering
\setlength{\tabcolsep}{3.0pt}
\renewcommand{\arraystretch}{1.15}
\caption*{Table 14. Qwen3-VL-8B per-class recall on the original T3 test set, and contribution to the macro-recall gap against ResNet-50 at 448 px.}
\begin{tabularx}{0.779\textwidth}{>{\hsize=0.767\hsize\RaggedRight\hyphenpenalty=10000\exhyphenpenalty=50\arraybackslash}X >{\hsize=0.754\hsize\RaggedRight\hyphenpenalty=10000\exhyphenpenalty=50\arraybackslash}X >{\hsize=0.443\hsize\RaggedRight\hyphenpenalty=10000\exhyphenpenalty=50\arraybackslash}X >{\hsize=2.036\hsize\RaggedRight\hyphenpenalty=10000\exhyphenpenalty=50\arraybackslash}X}
\toprule
Class & Test support & Recall & Contribution vs. ResNet-50 448 px \\
\midrule
corrosão & 184 & 96.20\% & +0.47 pt \\
good & 6,518 & 99.65\% & −0.03 pt \\
missing-cap & 136 & 74.26\% & +0.32 pt \\
nest & 103 & 93.20\% & +0.28 pt \\
peeling-paint & 4 & 75.00\% & +10.71 pt \\
rust & 334 & 80.84\% & −0.90 pt \\
torned-up & 5 & 0.00\% & −2.86 pt \\
\bottomrule
\end{tabularx}
\end{table}

The +7.99-point gap is dominated by the four-instance \texttt{peeling-paint} class, while ResNet-50 leads on \texttt{rust} and on the five-instance \texttt{torned-up} class. Nine test items across two classes account for +7.85 points of the +7.99-point total, which is a further reason not to treat the original-T3 difference as established.

Pixel-budget audit. To measure the effect of input budget directly, the processor \texttt{max\_pixels} value was constrained and the actual grid, visual-token count, input-identifier hash, pixel-tensor hash, prediction hash, and elapsed time were logged for every item. All six prediction files differ by SHA-256, and the lower budgets change the audited input relative to native resolution for 4,472, 3,727, 2,839, 2,025, and 1,063 of 7,284 items.

\begin{table}[htbp]
\centering
\setlength{\tabcolsep}{4pt}
\renewcommand{\arraystretch}{1.15}
\caption*{Table 15. Qwen3-VL-8B pixel-budget audit on the original T3 test set (n = 7,284). Macro recall uses the seven-class label space.}
\begin{tabularx}{\textwidth}{>{\hsize=1.273\hsize\RaggedRight\hyphenpenalty=10000\exhyphenpenalty=50\arraybackslash}X >{\hsize=0.914\hsize\RaggedRight\hyphenpenalty=10000\exhyphenpenalty=50\arraybackslash}X >{\hsize=0.798\hsize\RaggedRight\hyphenpenalty=10000\exhyphenpenalty=50\arraybackslash}X >{\hsize=0.871\hsize\RaggedRight\hyphenpenalty=10000\exhyphenpenalty=50\arraybackslash}X >{\hsize=0.797\hsize\RaggedRight\hyphenpenalty=10000\exhyphenpenalty=50\arraybackslash}X >{\hsize=1.347\hsize\RaggedRight\hyphenpenalty=10000\exhyphenpenalty=50\arraybackslash}X}
\toprule
\texttt{max\_pixels} & Inputs changed vs. native & Mean visual tokens & Median visual tokens & Macro recall & Prediction SHA-256 (prefix) \\
\midrule
65,536 & 4,472 & 62.7 & 64 & 0.7861 & \texttt{e14507e15251} \\
131,072 & 3,727 & 97.6 & 121 & 0.8261 & \texttt{a163d5c9a9e6} \\
200,704 & 2,839 & 129.7 & 144 & 0.8343 & \texttt{033dd1a4bf96} \\
401,408 & 2,025 & 189.3 & 144 & 0.8908 & \texttt{b22525e9c05b} \\
802,816 & 1,063 & 276.9 & 144 & 0.8926 & \texttt{ff9cc972daa2} \\
native (100,000,000) & 0 & 329.5 & 144 & 0.8939 & \texttt{03380898f7c5} \\
\bottomrule
\end{tabularx}
\end{table}

Reducing the budget from native to 65,536 pixels lowers Qwen3-VL-8B macro recall by 10.78 points. The two largest increments occur from 65,536 to 131,072 pixels, worth 4.01 points, and from 200,704 to 401,408 pixels, worth 5.66 points. Above 401,408 pixels the change is 0.31 points despite a further increase in mean visual tokens from 189.3 to 329.5. The median remains 144 tokens because many crops already fall below the higher budgets.

To test whether pixel budget accounts for the difference between the two VLMs, InternVL3.5-8B was evaluated under the same source-pixel budgets. This is a source-pixel control rather than a visual-token match, because InternVL's dynamic tiling is aspect-ratio driven and its patch and token counts do not track source area in the same way as the Qwen processor. The control is effective: the 65,536-pixel setting changes 4,699 InternVL pixel tensors relative to native resolution.

\begin{table}[htbp]
\centering
\setlength{\tabcolsep}{4pt}
\renewcommand{\arraystretch}{1.15}
\caption*{Table 16. Source-pixel-matched comparison of Qwen3-VL-8B and InternVL3.5-8B on the original T3 test set (n = 7,284). Inputs changed and patch and token statistics refer to the InternVL arm; Qwen token statistics are in Table 15.}
\begin{tabularx}{\textwidth}{>{\hsize=1.250\hsize\RaggedRight\hyphenpenalty=10000\exhyphenpenalty=50\arraybackslash}X >{\hsize=0.684\hsize\RaggedRight\hyphenpenalty=10000\exhyphenpenalty=50\arraybackslash}X >{\hsize=0.903\hsize\RaggedRight\hyphenpenalty=10000\exhyphenpenalty=50\arraybackslash}X >{\hsize=1.454\hsize\RaggedRight\hyphenpenalty=10000\exhyphenpenalty=50\arraybackslash}X >{\hsize=0.903\hsize\RaggedRight\hyphenpenalty=10000\exhyphenpenalty=50\arraybackslash}X >{\hsize=0.903\hsize\RaggedRight\hyphenpenalty=10000\exhyphenpenalty=50\arraybackslash}X >{\hsize=0.903\hsize\RaggedRight\hyphenpenalty=10000\exhyphenpenalty=50\arraybackslash}X}
\toprule
\texttt{max\_pixels} & Qwen macro recall & InternVL macro recall & Qwen–InternVL gap & InternVL inputs changed & InternVL mean patches & InternVL mean visual tokens \\
\midrule
65,536 & 0.7861 & 0.7118 & +7.43 pt & 4,699 & 1.35 & 346.1 \\
131,072 & 0.8261 & 0.7519 & +7.42 pt & 3,753 & 1.35 & 346.1 \\
200,704 & 0.8343 & 0.7566 & +7.77 pt & 2,976 & 1.35 & 346.1 \\
401,408 & 0.8908 & 0.7548 & +13.61 pt & 1,982 & 1.35 & 346.1 \\
802,816 & 0.8926 & 0.7710 & +12.16 pt & 1,092 & 2.36 & 605.1 \\
native (100,000,000) & 0.8939 & 0.7734 & +12.05 pt & 0 & 2.94 & 752.9 \\
\bottomrule
\end{tabularx}
\end{table}

Three patterns stand out. InternVL is less sensitive to source-pixel budget than Qwen, with a range of 6.16 points against 10.78 points. InternVL remains below Qwen at every budget, by 7.43 to 7.77 points at the three lowest budgets and 12.05 to 13.61 points at the three highest, so source-pixel budget alone does not account for the difference. The difference is also not a visual-token effect; it runs in the opposite direction. At native settings InternVL consumes a mean of 752.9 visual tokens per item against 329.5 for Qwen, so Qwen scores 12.05 points higher while consuming 56\% fewer visual tokens. An account in which the stronger model simply resolves more of the image is inconsistent with the measured token budgets. The experiment does not isolate the mechanism; tokenizer design, patch-formation policy, pre-training composition, and architecture-specific visual encoding remain candidate explanations, and separating them requires a matched-token comparison that this study does not contain. Figure 4 shows the budget curves and the token relationship.

\begin{figure}[!htbp]
\centering
\IfFileExists{figures/figure4.jpg}{\includegraphics[width=0.95\linewidth]{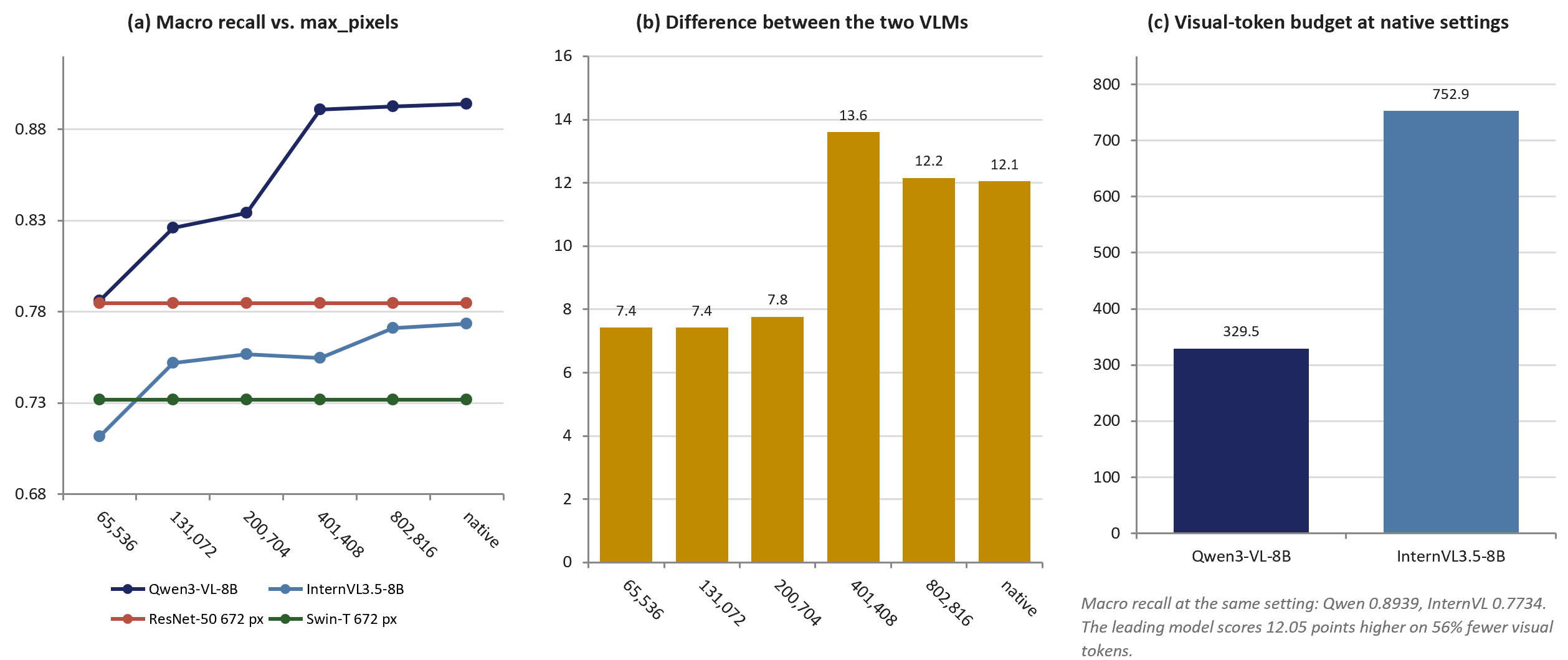}}{\fbox{\parbox{0.92\linewidth}{\centering\vspace{2.5cm}\small Placeholder for \texttt{figures/figure4.jpg}\vspace{2.5cm}}}}
\caption*{Figure 4. Resolution and pixel-budget audit. Panel (a) macro recall against input resolution for the vision backbones and against \texttt{max\_pixels} for both VLMs; panel (b) the Qwen–InternVL difference against the same budget axis; panel (c) macro recall against mean visual tokens for both VLMs, which shows Qwen achieving higher recall at lower token count. Data from Tables 10, 15, and 16.}
\end{figure}
\FloatBarrier

One consequence is that ResNet-50 at 448 px exceeds InternVL3.5-8B at every tested resolution but falls 8.00 points short of Qwen3-VL-8B at the vision backbone's highest-scoring setting. This is an effect size rather than an established advantage: the tower-clustered test gives a 95\% interval of −4.59 to +15.01 points and a Holm-adjusted p of 0.834, and Table 14 shows nine test items carrying almost the whole difference.

A complementary control constrains InternVL3.5-8B to a single 448 px tile, which lowers its balanced accuracy to 74.70\%. The drop from the dynamic-tiling baseline is directionally expected but not significant (b = 27, c = 18, p = 0.235). This control and the source-pixel control of Table 16 impose the constraint at different points in the preprocessing pipeline and are not two measurements of the same quantity.

\subsubsection*{5.7.2. Per-Class Decomposition}
\begin{table}[htbp]
\centering
\setlength{\tabcolsep}{3.0pt}
\renewcommand{\arraystretch}{1.15}
\caption*{Table 17. Per-class recall (\%) at K = 7 on the T3 test set: InternVL3.5-8B + LoRA at native tiling against Swin-T at 448 px. The 448 px setting is the highest-scoring point of the vision-only sweep rather than the matched budget of approximately 768 px.}
\begin{tabularx}{\textwidth}{>{\hsize=0.877\hsize\RaggedRight\hyphenpenalty=10000\exhyphenpenalty=50\arraybackslash}X >{\hsize=0.774\hsize\RaggedRight\hyphenpenalty=10000\exhyphenpenalty=50\arraybackslash}X >{\hsize=0.692\hsize\RaggedRight\hyphenpenalty=10000\exhyphenpenalty=50\arraybackslash}X >{\hsize=1.050\hsize\RaggedRight\hyphenpenalty=10000\exhyphenpenalty=50\arraybackslash}X >{\hsize=1.607\hsize\RaggedRight\hyphenpenalty=10000\exhyphenpenalty=50\arraybackslash}X}
\toprule
Class & Test support & VLM recall & Vision-only recall & Contribution to macro gap \\
\midrule
corrosão & 184 & 90.76 & 93.48 & −0.39 \\
good & 6,518 & 99.82 & 99.83 & −0.00 \\
missing-cap & 136 & 69.85 & 88.24 & −2.63 \\
nest & 103 & 98.06 & 98.06 & 0.00 \\
peeling-paint & 4 & 100.00 & 0.00 & +14.29 \\
rust & 334 & 86.83 & 91.62 & −0.68 \\
torned-up & 5 & 0.00 & 60.00 & −8.57 \\
Macro-average & 7,284 & 77.90 & 75.89 & +2.01 \\
\bottomrule
\end{tabularx}
\end{table}

The decomposition is internally consistent: the seven per-class contributions sum to the 2.01-point macro gap, and both column means reproduce their macro-averages. Because an unweighted macro-average gives each class one-seventh of the score, a four-instance class carries the same weight as a class with thousands of instances, so the decomposition is intrinsically sensitive to rare-class sampling. The entire seven-way advantage of this VLM over this vision-only run resides in \texttt{peeling-paint}. On the other six classes the vision backbone leads by 12.3 points in aggregate, most substantially on \texttt{missing-cap}, with 136 test instances at 88.2\% against 69.8\%, and on \texttt{torned-up}, with 5 test instances at 60.0\% against 0.0\%.

Three scope conditions apply. The vision-only column is Swin-T at the highest-scoring point of its sweep rather than at the matched budget; at the bracketing 672 px setting Swin-T scores 73.20\% and the gap widens to +4.70 points. Swin-T is also the weaker vision-only backbone at this granularity, since ResNet-50 reaches 81.08\% at 448 px and 78.47\% at 672 px, so comparing against Swin-T favours the VLM; against ResNet-50 at 448 px the same InternVL model is 3.18 points behind. Finally, the decomposition covers InternVL3.5-8B, while the corresponding decomposition for Qwen3-VL-8B is Table 14, which shows the same concentration in rare classes in a different configuration.

What carries over is the mechanism, not the exact residual. On a seven-class macro-average whose two rarest classes hold nine test items between them, those nine items can move the headline number by 22.9 points in opposite directions. This holds regardless of which pair of models is compared and is why the decomposition is reported alongside every macro-average in this paper. Figure 5 shows the split-level version of the same decomposition.

\begin{figure}[!htbp]
\centering
\IfFileExists{figures/figure5.jpg}{\includegraphics[width=0.95\linewidth]{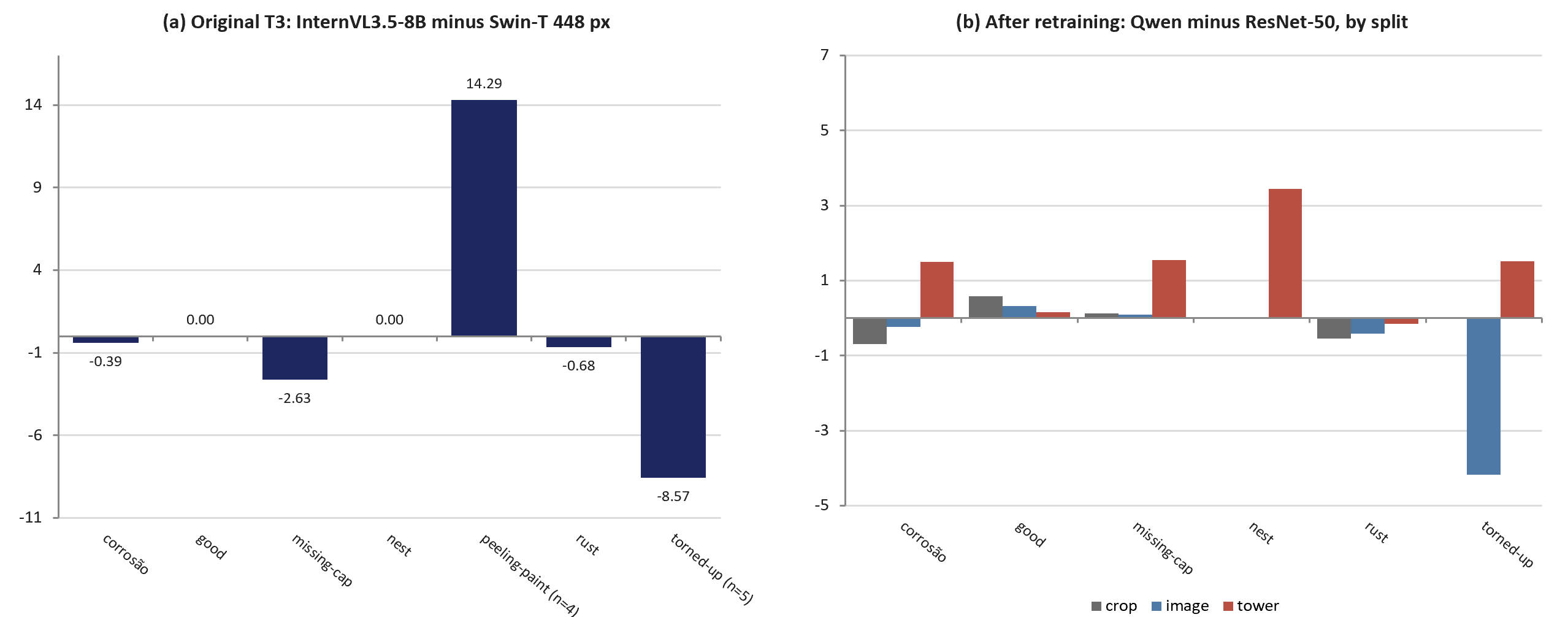}}{\fbox{\parbox{0.92\linewidth}{\centering\vspace{2.5cm}\small Placeholder for \texttt{figures/figure5.jpg}\vspace{2.5cm}}}}
\caption*{Figure 5. Per-class contributions to the macro-recall difference. Panel (a) the original T3 comparison of Table 17; panel (b) the crop, image, and tower contributions of Table 25 with their randomization intervals. Both panels annotate \texttt{peeling-paint} and \texttt{torned-up} to show their weight relative to their test support.}
\end{figure}
\FloatBarrier

\subsubsection*{5.7.3. Prompt Ablation}
\begin{table}[htbp]
\centering
\setlength{\tabcolsep}{3.0pt}
\renewcommand{\arraystretch}{1.15}
\caption*{Table 18. Prompt ablation, InternVL3.5-8B + LoRA, T3 test (n = 7,284), K = 7. P3 supplies the asset description, decision tree, and candidate options; P1 supplies the question stem, asset name, and candidate options only.}
\begin{tabularx}{0.667\textwidth}{>{\hsize=1.159\hsize\RaggedRight\hyphenpenalty=10000\exhyphenpenalty=50\arraybackslash}X >{\hsize=1.198\hsize\RaggedRight\hyphenpenalty=10000\exhyphenpenalty=50\arraybackslash}X >{\hsize=0.527\hsize\RaggedRight\hyphenpenalty=10000\exhyphenpenalty=50\arraybackslash}X >{\hsize=1.116\hsize\RaggedRight\hyphenpenalty=10000\exhyphenpenalty=50\arraybackslash}X}
\toprule
Training prompt & Inference prompt & BalAcc & Δ from baseline \\
\midrule
P3 & P3 (baseline) & 77.90\% & — \\
P3 & P1 & 14.34\% & −63.56 \\
P1 & P1 (controlled) & 14.29\% & −63.62 \\
P1 & P3 (cross) & 69.26\% & −8.64 \\
\bottomrule
\end{tabularx}
\end{table}

Removing the definitional side information collapses the model to 14.3\%, and the controlled P1-trained arm reproduces that collapse to within 0.05 points, which excludes train–test prompt mismatch as the cause. The collapse mechanism is systematic majority-class prediction rather than random guessing: without the decision tree the model labels essentially every item \texttt{good}, giving a macro recall of one correct class in seven. The cross arm separates the definitional contribution into an inference-time channel worth 54.97 points and a training-time channel worth 8.64 points.

Language grounding is therefore a necessary condition for these VLMs to perform fine-grained typing at all. It does not distinguish one VLM from another: the ablation was replicated on Qwen3-VL-8B, and the bare-prompt collapse is identical (Table 19). Because the decision tree is itself side information, this result describes the full VLM system, image plus prompt, rather than showing that the vision-only backbone received an equivalent input.

\begin{table}[htbp]
\centering
\setlength{\tabcolsep}{3.0pt}
\renewcommand{\arraystretch}{1.15}
\caption*{Table 19. Prompt ablation replicated on Qwen3-VL-8B, T3 test (n = 7,284), K = 7.}
\begin{tabularx}{\textwidth}{>{\hsize=1.246\hsize\RaggedRight\hyphenpenalty=10000\exhyphenpenalty=50\arraybackslash}X >{\hsize=0.900\hsize\RaggedRight\hyphenpenalty=10000\exhyphenpenalty=50\arraybackslash}X >{\hsize=0.854\hsize\RaggedRight\hyphenpenalty=10000\exhyphenpenalty=50\arraybackslash}X}
\toprule
Configuration & InternVL3.5-8B + LoRA & Qwen3-VL-8B + LoRA \\
\midrule
P3 baseline & 77.90\% & 89.08\% \\
P1 + P1 (controlled) & 14.29\% & 14.29\% \\
P1 + P3 (cross) & 69.26\% & 84.36\% \\
Inference-time definitional channel & 54.97 pt & 70.07 pt \\
Training-time definitional channel & 8.64 pt & 4.72 pt \\
\bottomrule
\end{tabularx}
\end{table}

Both models collapse to the identical 14.29\%, the value expected from majority-class prediction under a seven-class macro average, which is independent of model identity. What differs is recovery. Qwen3-VL-8B regains 70.07 points from the bare-prompt floor when the decision tree is restored at inference time without further training, against 54.97 points for InternVL3.5-8B, and correspondingly relies less on the training-time channel, 4.72 points against 8.64. Read together with Table 16, this describes a model that extracts more from the same textual side information while consuming fewer visual tokens. That is a description of the difference between the two VLMs rather than an explanation of it, since neither experiment isolates whether the advantage originates in the language tower, the vision tower, the tokenizer, or the pre-training corpus. Figure 6 shows the four prompt arms for both models.

\begin{figure}[!htbp]
\centering
\IfFileExists{figures/figure6.jpg}{\includegraphics[width=0.95\linewidth]{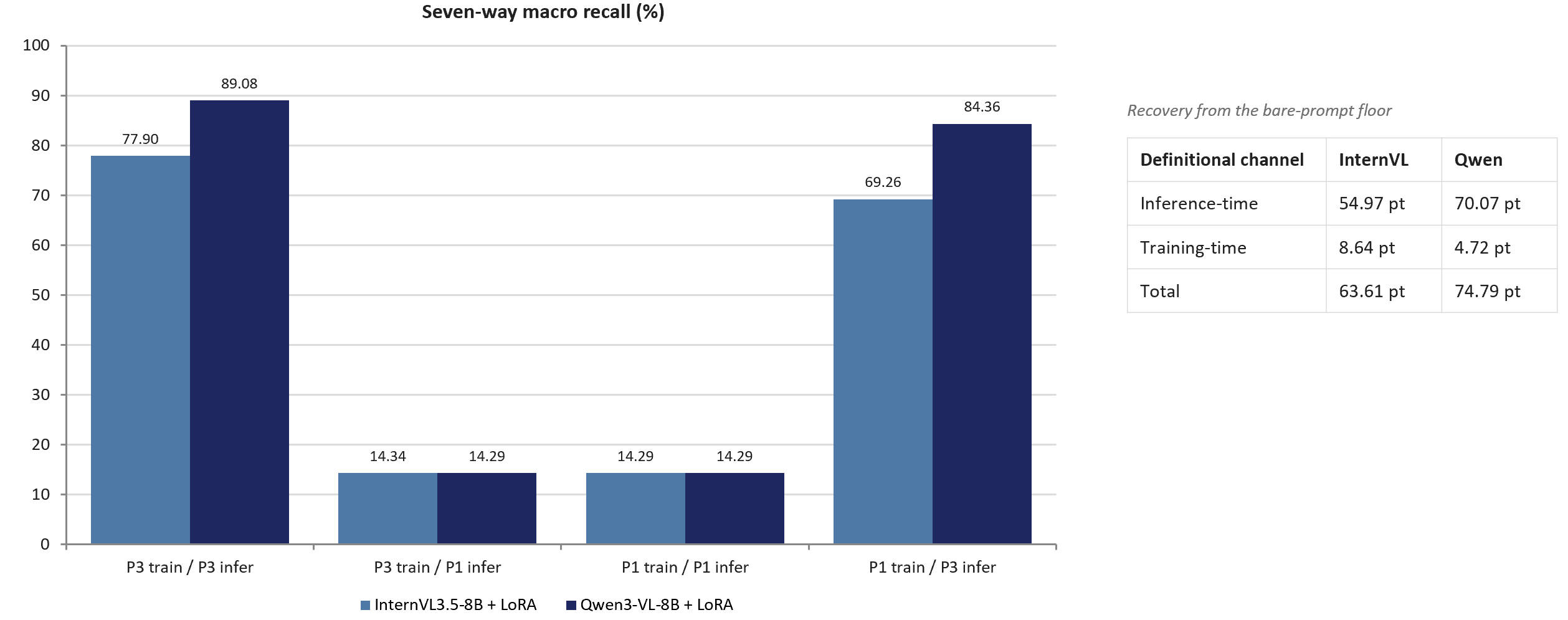}}{\fbox{\parbox{0.92\linewidth}{\centering\vspace{2.5cm}\small Placeholder for \texttt{figures/figure6.jpg}\vspace{2.5cm}}}}
\caption*{Figure 6. Prompt ablation across the four training and inference combinations for both adapted VLMs, from Tables 18 and 19, with the inference-time and training-time channels annotated.}
\end{figure}
\FloatBarrier

\subsubsection*{5.7.4. Summary of the Controlled Factors}
\begin{table}[htbp]
\centering
\setlength{\tabcolsep}{4pt}
\renewcommand{\arraystretch}{1.15}
\caption*{Table 20. Controlled factors for the seven-way comparison and the verdict each experiment supports.}
\begin{tabularx}{\textwidth}{>{\hsize=0.891\hsize\RaggedRight\hyphenpenalty=10000\exhyphenpenalty=50\arraybackslash}X >{\hsize=1.008\hsize\RaggedRight\hyphenpenalty=10000\exhyphenpenalty=50\arraybackslash}X >{\hsize=1.064\hsize\RaggedRight\hyphenpenalty=10000\exhyphenpenalty=50\arraybackslash}X >{\hsize=1.037\hsize\RaggedRight\hyphenpenalty=10000\exhyphenpenalty=50\arraybackslash}X}
\toprule
Factor & Controlling experiment & Effect & Role \\
\midrule
Input pixel budget, vision-only side & 224 px to bracketing settings, same backbone (Table 11) & ResNet-50 gap to InternVL: +20.53 to −0.57 pt at 672 px; Swin-T gap: +23.67 to +4.70 pt & Primary for the vision-only versus InternVL comparison \\
Source-pixel budget, VLM side & \texttt{max\_pixels} audit, both VLMs (Tables 15, 16) & Qwen varies 10.78 pt, InternVL 6.16 pt; Qwen ahead at every matched budget & Does not account for the Qwen–InternVL difference \\
Visual-token budget & Token counts logged in the same audit & Qwen leads by 12.05 pt at native while using 329.5 against 752.9 mean tokens & Excluded; the difference runs opposite to a token-budget account \\
Macro-averaging over rare classes & Per-class decompositions (Tables 14, 17) & Nine test items across two classes move the macro-average by up to 22.9 pt & Metric artefact; magnitude depends on configuration \\
Language grounding (side information) & P3 to P1, both VLMs (Tables 18, 19) & Both collapse to an identical 14.29\% without the decision tree & Necessary for every VLM tested; not what distinguishes them \\
Model scale within a family & InternVL 2B to 8B (Tables 6, 10) & +0.64 pt at K = 2 (p = 0.053); +3.74 pt at K = 7 & Small \\
Coarse architecture & Swin-T is a transformer and trails ResNet-50 at K = 7 & — & Convolutional versus transformer excluded; architecture-specific encoding not excluded \\
Split construction and seed & Crop, image, and tower retraining and tower14 replication (Tables 23, 27) & Sign of the Qwen–ResNet difference reverses across splits and seeds & Decisive; no stable family ordering \\
Model identity & Residual after all of the above & Qwen–InternVL: +11.18 pt at K = 7, +1.3 pt at K = 2 & Bounded but not explained \\
\bottomrule
\end{tabularx}
\end{table}

The final row is the one factor that the controls bound without explaining. Two 8B-parameter adapted VLMs, trained on the same corpus with the same hyperparameters and scored on the same partition, differ by 1.3 points at binary screening and by 11.18 points at seven-way typing at native settings. The pixel-budget audit establishes that Qwen is sensitive to its input budget, excludes a token-budget account in the only direction such an account could take, and leaves the difference intact at every matched source-pixel setting. Tokenizer design, patch-formation policy, pre-training composition, and architecture-specific visual encoding therefore remain candidate explanations. A matched-visual-token comparison is the experiment that would separate them.

\subsection*{5.8. Quantization and Deployment Envelope}
\begin{table}[htbp]
\centering
\setlength{\tabcolsep}{3.0pt}
\renewcommand{\arraystretch}{1.15}
\caption*{Table 21. Inference precision, InternVL3.5-8B + LoRA, full test sets (T2 n = 2,021; T3 n = 7,284), with paired McNemar tests against bf16 and throughput measured over 50 items.}
\begin{tabularx}{\textwidth}{>{\hsize=1.028\hsize\RaggedRight\hyphenpenalty=10000\exhyphenpenalty=50\arraybackslash}X >{\hsize=1.198\hsize\RaggedRight\hyphenpenalty=10000\exhyphenpenalty=50\arraybackslash}X >{\hsize=0.692\hsize\RaggedRight\hyphenpenalty=10000\exhyphenpenalty=50\arraybackslash}X >{\hsize=1.198\hsize\RaggedRight\hyphenpenalty=10000\exhyphenpenalty=50\arraybackslash}X >{\hsize=1.363\hsize\RaggedRight\hyphenpenalty=10000\exhyphenpenalty=50\arraybackslash}X >{\hsize=0.941\hsize\RaggedRight\hyphenpenalty=10000\exhyphenpenalty=50\arraybackslash}X >{\hsize=1.358\hsize\RaggedRight\hyphenpenalty=10000\exhyphenpenalty=50\arraybackslash}X >{\hsize=0.611\hsize\RaggedRight\hyphenpenalty=10000\exhyphenpenalty=50\arraybackslash}X >{\hsize=0.611\hsize\RaggedRight\hyphenpenalty=10000\exhyphenpenalty=50\arraybackslash}X}
\toprule
Precision & T2 BalAcc & T2 β & T3 BalAcc & T3 macro-F1 & Memory & Throughput & T2 p & T3 p \\
\midrule
bf16 & 0.9586 & 7.57\% & 0.7790 & 0.7954 & ≈16 GB & 1.54 img/s & — & — \\
INT8 & 0.9592 & 7.44\% & 0.7790 & 0.7950 & ≈9 GB & 1.17 img/s & 1.000 & 0.625 \\
NF4 & 0.9592 & 7.44\% & 0.7806 & 0.7964 & 6.2 GB & 1.49 img/s & 1.000 & 1.000 \\
\bottomrule
\end{tabularx}
\end{table}

No precision differs from bf16 by more than 0.16 points on either test set, all p-values exceed 0.6, and the three to five discordant pairs per comparison are consistent with sampling noise. These results do not support a material accuracy difference at the observed resolution, but no formal equivalence margin was pre-specified, so the statement is descriptive rather than a powered equivalence claim. The near-identical T3 scores do not imply identical per-item predictions, since the discordant pairs disagree in both directions. FP8 quantization failed with a meta-tensor materialization error specific to this architecture under the evaluated framework versions and is treated as a tooling limitation rather than a property of FP8. These results cover InternVL3.5-8B only and are not extrapolated to other VLMs.

The deployment consequences differ between the two low-precision modes. NF4 reduces memory by 61\% at a 3\% throughput cost, whereas INT8 reduces memory by 44\% at a 24\% throughput cost arising from dequantization overhead. On this single-GPU, 50-item measurement NF4 is the preferable operating point on both measured axes, which is not a general deployment recommendation. All five candidate backbones load and accept LoRA injection within a single 96 GB card, with bf16 inference footprints ranging from 4.0 to 4.4 GB at the 2B tier and 15.5 to 16.3 GB at the 7B to 8B tier.

A separate vision-only benchmark measures trained ResNet-50 and Swin-T checkpoints at 448 and 672 px using 512 test images, five repeats per setting, 200 single-image latency samples, and batch sizes from 1 to 64. Latency includes the host-to-device copy and the forward pass, and throughput also includes host-to-device transfer. The best mean batch throughput occurs at batch size 32 for every configuration.

\begin{table}[htbp]
\centering
\setlength{\tabcolsep}{4pt}
\renewcommand{\arraystretch}{1.15}
\caption*{Table 22. Vision-only deployment benchmark on one NVIDIA RTX PRO 6000 Blackwell (PyTorch 2.11.0+cu128; 512 images; five repeats; latency n = 200). Peak memory is measured over the full batch sweep rather than model weights alone.}
\begin{tabularx}{\textwidth}{>{\hsize=0.569\hsize\RaggedRight\hyphenpenalty=10000\exhyphenpenalty=50\arraybackslash}X >{\hsize=0.866\hsize\RaggedRight\hyphenpenalty=10000\exhyphenpenalty=50\arraybackslash}X >{\hsize=1.037\hsize\RaggedRight\hyphenpenalty=10000\exhyphenpenalty=50\arraybackslash}X >{\hsize=0.964\hsize\RaggedRight\hyphenpenalty=10000\exhyphenpenalty=50\arraybackslash}X >{\hsize=0.964\hsize\RaggedRight\hyphenpenalty=10000\exhyphenpenalty=50\arraybackslash}X >{\hsize=1.300\hsize\RaggedRight\hyphenpenalty=10000\exhyphenpenalty=50\arraybackslash}X >{\hsize=1.300\hsize\RaggedRight\hyphenpenalty=10000\exhyphenpenalty=50\arraybackslash}X}
\toprule
Input & Backbone & Peak memory & p50 latency & p95 latency & Best batch throughput & Sustained throughput (batch 64) \\
\midrule
448 px & ResNet-50 & 5.97 GB & 4.49 ms & 4.63 ms & 806.4 img/s (batch 32) & 747.8 img/s \\
448 px & Swin-T & 3.70 GB & 6.28 ms & 6.36 ms & 525.7 img/s (batch 32) & 524.7 img/s \\
672 px & ResNet-50 & 22.03 GB & 6.29 ms & 6.46 ms & 344.8 img/s (batch 32) & 340.8 img/s \\
672 px & Swin-T & 8.19 GB & 6.63 ms & 6.75 ms & 241.6 img/s (batch 32) & 234.0 img/s \\
\bottomrule
\end{tabularx}
\end{table}

At 448 px ResNet-50 has the highest throughput and Swin-T the lowest memory. At 672 px the peak memory of ResNet-50 rises to 22.03 GB, an activation-memory effect at large batch sizes rather than a model-weight comparison, while Swin-T remains below 10 GB. These are workstation measurements with preloaded tensors and do not include JPEG decoding, disk input, energy, thermal throttling, or an embedded accelerator. The VLM throughput in Table 21 and the vision-only throughput in Table 22 were measured with different harnesses and batch policies and are reported separately rather than merged into a single ranking. Figure 7 shows the three deployment axes.

\begin{figure}[!htbp]
\centering
\IfFileExists{figures/figure7.jpg}{\includegraphics[width=0.95\linewidth]{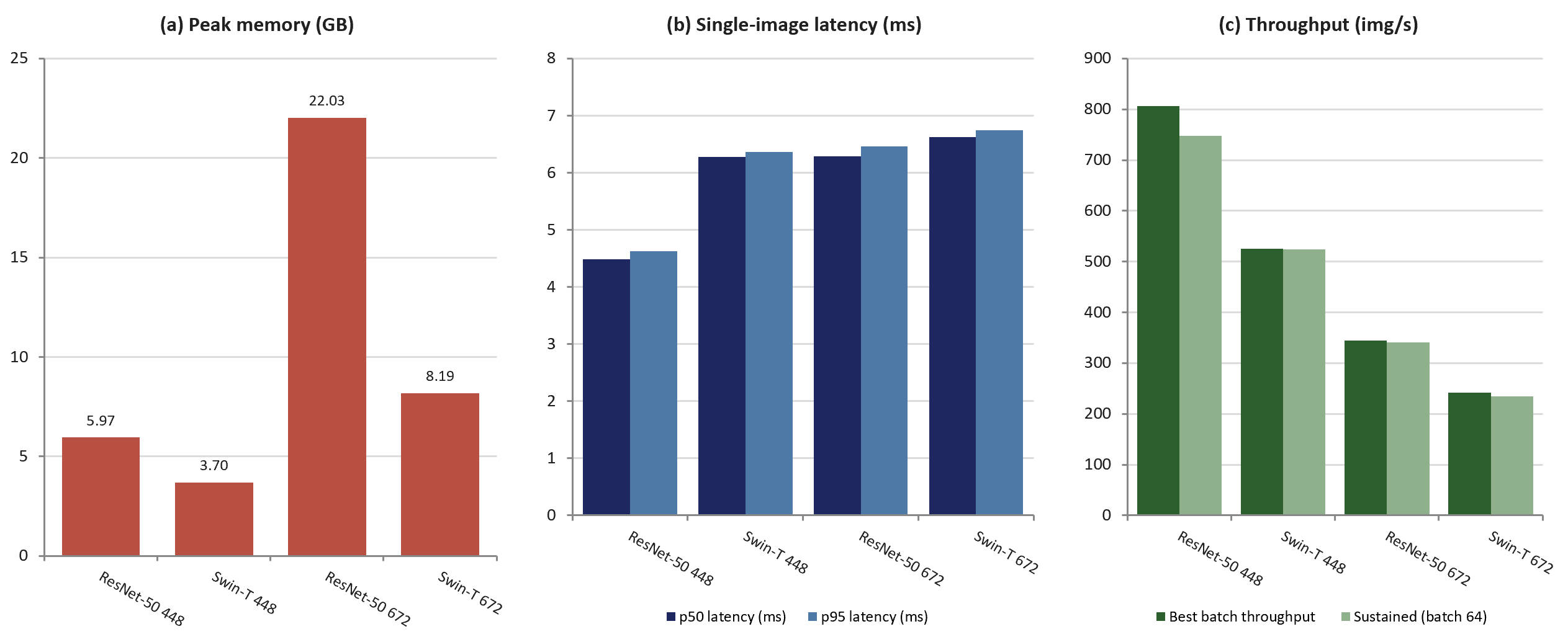}}{\fbox{\parbox{0.92\linewidth}{\centering\vspace{2.5cm}\small Placeholder for \texttt{figures/figure7.jpg}\vspace{2.5cm}}}}
\caption*{Figure 7. Vision-only deployment benchmark. Peak memory, p50 and p95 latency, and sustained throughput for ResNet-50 and Swin-T at 448 and 672 px, from Table 22, in three panels grouped by backbone and resolution.}
\end{figure}
\FloatBarrier

\subsection*{5.9. Split-Level Retraining and Tower Replication}
A checkpoint-provenance audit showed that the original crop-level and image-level rows reused checkpoints that had previously seen about 70\% of the new test items. Qwen3-VL-8B and ResNet-50 at 448 px were therefore retrained on the crop-level and image-level splits; the tower-level models were already trained on the tower-disjoint split. The three regimes differ in residual dependence:

\begin{enumerate}
\item Crop-level split. Retraining removes exact test-item leakage, but same-image and same-tower overlap remains by design: 488 of 6,672 test image paths also occur in the training file, and all 42 test towers occur in training.
\item Image-level split. No image path or \texttt{parent\_id} crosses train and test, but all 41 test towers occur in training.
\item Tower-disjoint split. No image path, parent group, or tower crosses train and test.
\end{enumerate}

All primary comparisons use the six classes common to the three test files: \texttt{corrosão}, \texttt{good}, \texttt{missing-cap}, \texttt{nest}, \texttt{rust}, and \texttt{torned-up}. Randomization is cluster-aware, with \texttt{\_tower} as the cluster unit for the crop and tower splits and \texttt{parent\_id} for the image split, and Holm correction is applied across the three split contrasts.

\begin{table}[htbp]
\centering
\setlength{\tabcolsep}{4pt}
\renewcommand{\arraystretch}{1.15}
\caption*{Table 23. Common-six-class macro recall after split-specific retraining. Holm family: the three split contrasts. Intervals are cluster sign-flip randomization intervals derived from the same reference distribution as the p-values.}
\begin{tabularx}{\textwidth}{>{\hsize=0.591\hsize\RaggedRight\hyphenpenalty=10000\exhyphenpenalty=50\arraybackslash}X >{\hsize=1.192\hsize\RaggedRight\hyphenpenalty=10000\exhyphenpenalty=50\arraybackslash}X >{\hsize=1.232\hsize\RaggedRight\hyphenpenalty=10000\exhyphenpenalty=50\arraybackslash}X >{\hsize=0.978\hsize\RaggedRight\hyphenpenalty=10000\exhyphenpenalty=50\arraybackslash}X >{\hsize=1.199\hsize\RaggedRight\hyphenpenalty=10000\exhyphenpenalty=50\arraybackslash}X >{\hsize=0.651\hsize\RaggedRight\hyphenpenalty=10000\exhyphenpenalty=50\arraybackslash}X >{\hsize=0.741\hsize\RaggedRight\hyphenpenalty=10000\exhyphenpenalty=50\arraybackslash}X >{\hsize=1.416\hsize\RaggedRight\hyphenpenalty=10000\exhyphenpenalty=50\arraybackslash}X}
\toprule
Split & Qwen3-VL-8B & ResNet-50 448 px & Difference & 95\% CI & Raw p & Holm p & Verdict \\
\midrule
crop & 0.9413 & 0.9464 & −0.51 pt & [−6.12, +5.08] & 0.690 & 0.804 & not detected \\
image & 0.9227 & 0.9667 & −4.40 pt & [−9.99, +1.16] & 0.402 & 0.804 & not detected \\
tower & 0.7647 & 0.6845 & +8.02 pt & [+0.71, +15.33] & 0.0239 & 0.0717 & not detected after correction \\
\bottomrule
\end{tabularx}
\end{table}

The corresponding full-label-space image result is also negative: Qwen reaches 0.9337 and ResNet-50 0.9715, a difference of −3.77 points. The single \texttt{peeling-paint} test item is predicted correctly by both models and therefore does not create the rare-class artefact seen before retraining.

\begin{table}[htbp]
\centering
\setlength{\tabcolsep}{3.0pt}
\renewcommand{\arraystretch}{1.15}
\caption*{Table 24. Per-class recall after split-specific retraining, common six classes.}
\begin{tabularx}{\textwidth}{>{\hsize=0.972\hsize\RaggedRight\hyphenpenalty=10000\exhyphenpenalty=50\arraybackslash}X >{\hsize=0.904\hsize\RaggedRight\hyphenpenalty=10000\exhyphenpenalty=50\arraybackslash}X >{\hsize=1.031\hsize\RaggedRight\hyphenpenalty=10000\exhyphenpenalty=50\arraybackslash}X >{\hsize=0.988\hsize\RaggedRight\hyphenpenalty=10000\exhyphenpenalty=50\arraybackslash}X >{\hsize=1.059\hsize\RaggedRight\hyphenpenalty=10000\exhyphenpenalty=50\arraybackslash}X >{\hsize=0.988\hsize\RaggedRight\hyphenpenalty=10000\exhyphenpenalty=50\arraybackslash}X >{\hsize=1.058\hsize\RaggedRight\hyphenpenalty=10000\exhyphenpenalty=50\arraybackslash}X}
\toprule
Class & crop Qwen & crop ResNet & image Qwen & image ResNet & tower Qwen & tower ResNet \\
\midrule
corrosão & 0.9076 & 0.9496 & 0.9353 & 0.9496 & 0.9005 & 0.8104 \\
good & 0.9983 & 0.9627 & 0.9975 & 0.9784 & 0.9979 & 0.9886 \\
missing-cap & 0.9921 & 0.9843 & 0.9759 & 0.9699 & 0.9380 & 0.8450 \\
nest & 1.0000 & 1.0000 & 0.9710 & 0.9710 & 0.8238 & 0.6166 \\
rust & 0.9164 & 0.9486 & 0.9065 & 0.9315 & 0.8370 & 0.8463 \\
torned-up & 0.8333 & 0.8333 & 0.7500 & 1.0000 & 0.0909 & 0.0000 \\
\bottomrule
\end{tabularx}
\end{table}

\begin{table}[htbp]
\centering
\setlength{\tabcolsep}{4pt}
\renewcommand{\arraystretch}{1.15}
\caption*{Table 25. Cluster sign-flip randomization intervals for per-class contributions to the Qwen–ResNet macro-recall difference (percentage points), common six classes.}
\begin{tabularx}{\textwidth}{>{\hsize=0.525\hsize\RaggedRight\hyphenpenalty=10000\exhyphenpenalty=50\arraybackslash}X >{\hsize=1.159\hsize\RaggedRight\hyphenpenalty=10000\exhyphenpenalty=50\arraybackslash}X >{\hsize=1.158\hsize\RaggedRight\hyphenpenalty=10000\exhyphenpenalty=50\arraybackslash}X >{\hsize=1.158\hsize\RaggedRight\hyphenpenalty=10000\exhyphenpenalty=50\arraybackslash}X}
\toprule
Class & Crop contribution [95\% CI] & Image contribution [95\% CI] & Tower contribution [95\% CI] \\
\midrule
corrosão & −0.70 [−1.68, +0.28] & −0.24 [−0.96, +0.48] & +1.50 [−0.47, +3.48] \\
good & +0.59 [+0.25, +0.93] & +0.32 [+0.25, +0.38] & +0.16 [+0.01, +0.30] \\
missing-cap & +0.13 [−0.26, +0.52] & +0.10 [−0.80, +1.00] & +1.55 [+0.00, +3.10] \\
nest & +0.00 [+0.00, +0.00] & +0.00 [−0.97, +0.97] & +3.45 [+0.00, +6.91] \\
rust & −0.54 [−1.18, +0.11] & −0.42 [−0.93, +0.10] & −0.15 [−0.62, +0.31] \\
torned-up & +0.00 [−5.56, +5.56] & −4.17 [−8.33, +0.00] & +1.52 [+0.00, +3.03] \\
\bottomrule
\end{tabularx}
\end{table}

The per-class contributions sum to the macro-recall differences of Table 23 in all three regimes, giving −0.52, −4.41, and +8.03 points against the tabulated −0.51, −4.40, and +8.02. The composition differs by regime. At image level the difference is carried almost entirely by \texttt{torned-up}, a class with a handful of test instances. At tower level it is distributed across \texttt{nest}, \texttt{missing-cap}, \texttt{torned-up}, and \texttt{corrosão}, with \texttt{rust} contributing −0.15 points, so the tower advantage is not concentrated in a single rare class in the way the original-T3 comparison of Table 14 is. The three regimes have different test-set sizes because they are separate split diagnostics rather than a single repeated-measures design, and n is reported for transparency rather than as evidence that the regimes are comparable in sample size.

Three conclusions follow. The earlier Qwen advantages at crop and image level were products of checkpoint reuse: after retraining, Qwen is slightly behind at crop level and 4.40 points behind at image level, with intervals that include zero. The tower-disjoint split retains a +8.02-point Qwen advantage with a raw cluster-aware p of 0.0239, which does not survive Holm correction across the three split comparisons. No split regime provides corrected evidence that Qwen3-VL-8B outperforms ResNet-50 under the common-six-class protocol.

Seed sensitivity on the nine-tower split. Adding Qwen seed 43 and ResNet-50 seeds 43 and 44 to the original seed-42 runs leaves Qwen ahead in all three contrasts, but the same-seed effect falls from +8.02 to +3.29 points and the seed-43 interval includes zero.

\begin{table}[htbp]
\centering
\setlength{\tabcolsep}{4pt}
\renewcommand{\arraystretch}{1.15}
\caption*{Table 26. Tower-disjoint multi-seed sensitivity analysis (common six classes; n = 13,574; nine test towers). Holm family: the three contrasts in this table. The same seed-42 contrast carries Holm p = 0.0717 in Table 23 because that table's family is the three split regimes.}
\begin{tabularx}{\textwidth}{>{\hsize=1.538\hsize\RaggedRight\hyphenpenalty=10000\exhyphenpenalty=50\arraybackslash}X >{\hsize=1.085\hsize\RaggedRight\hyphenpenalty=10000\exhyphenpenalty=50\arraybackslash}X >{\hsize=1.155\hsize\RaggedRight\hyphenpenalty=10000\exhyphenpenalty=50\arraybackslash}X >{\hsize=0.840\hsize\RaggedRight\hyphenpenalty=10000\exhyphenpenalty=50\arraybackslash}X >{\hsize=1.188\hsize\RaggedRight\hyphenpenalty=10000\exhyphenpenalty=50\arraybackslash}X >{\hsize=0.558\hsize\RaggedRight\hyphenpenalty=10000\exhyphenpenalty=50\arraybackslash}X >{\hsize=0.636\hsize\RaggedRight\hyphenpenalty=10000\exhyphenpenalty=50\arraybackslash}X}
\toprule
Comparison & Qwen macro recall & ResNet macro recall & Difference & 95\% CI & Raw p & Holm p \\
\midrule
seed 42 vs. seed 42 & 0.7647 & 0.6845 & +8.02 pt & [+0.71, +15.33] & 0.0239 & 0.0478 \\
seed 43 vs. seed 43 & 0.7560 & 0.7231 & +3.29 pt & [−2.97, +9.55] & 0.4499 & 0.4499 \\
seed 43 vs. seed 44 (sensitivity) & 0.7560 & 0.7057 & +5.03 pt & [+0.41, +9.66] & 0.0091 & 0.0273 \\
\bottomrule
\end{tabularx}
\end{table}

The third row pairs Qwen seed 43 against ResNet seed 44 and is a sensitivity contrast rather than a matched-seed replication. Across the available runs Qwen averages 0.7604 with a sample standard deviation of 0.0061 over two seeds and ResNet-50 averages 0.7044 with 0.0194 over three seeds. The mean seed-matched difference is +5.66 points with a range of 4.73 points, so the direction is consistent but the magnitude and inferential status are seed-sensitive.

A larger tower replication. Because the nine-tower analysis used few test towers and an unbalanced seed design, a larger replication was constructed with 18 training towers, 10 validation towers, and 14 test towers. Its T3-only test set contains 5,536 items and both families are run with seeds 42, 43, and 44. This addresses the two principal limitations of Table 26, although it remains a new split rather than a pre-registered confirmatory trial.

\begin{table}[htbp]
\centering
\setlength{\tabcolsep}{3.0pt}
\renewcommand{\arraystretch}{1.15}
\caption*{Table 27. Tower14 replication: common-six-class macro recall and tower-clustered inference (n = 5,536; 14 test towers; three matched seeds per family). Holm family: the three seed contrasts in this table.}
\begin{tabularx}{\textwidth}{>{\hsize=0.446\hsize\RaggedRight\hyphenpenalty=10000\exhyphenpenalty=50\arraybackslash}X >{\hsize=1.302\hsize\RaggedRight\hyphenpenalty=10000\exhyphenpenalty=50\arraybackslash}X >{\hsize=1.605\hsize\RaggedRight\hyphenpenalty=10000\exhyphenpenalty=50\arraybackslash}X >{\hsize=0.921\hsize\RaggedRight\hyphenpenalty=10000\exhyphenpenalty=50\arraybackslash}X >{\hsize=1.415\hsize\RaggedRight\hyphenpenalty=10000\exhyphenpenalty=50\arraybackslash}X >{\hsize=0.613\hsize\RaggedRight\hyphenpenalty=10000\exhyphenpenalty=50\arraybackslash}X >{\hsize=0.698\hsize\RaggedRight\hyphenpenalty=10000\exhyphenpenalty=50\arraybackslash}X}
\toprule
Seed & Qwen3-VL-8B & ResNet-50 448 px & Difference & 95\% CI & Raw p & Holm p \\
\midrule
42 & 0.9369 & 0.9561 & −1.91 pt & [−3.96, +0.13] & 0.0904 & 0.2712 \\
43 & 0.9517 & 0.9455 & +0.62 pt & [−0.87, +2.10] & 0.9189 & 1.0000 \\
44 & 0.8370 & 0.9512 & −11.42 pt & [−24.04, +1.22] & 0.6482 & 1.0000 \\
\bottomrule
\end{tabularx}
\end{table}

Across the three seeds Qwen averages 0.9085 common-six macro recall with a sample standard deviation of 0.0624, and ResNet-50 averages 0.9509 with 0.0053. The mean difference is −4.24 points with a range of 12.04 points. ResNet-50 leads in two of three seeds, and the Qwen seed-44 run falls to 0.8370. Under the full seven-class label space Qwen averages 0.8740 against 0.8151 for ResNet-50, but the seed-44 contrast again reverses, at −2.65 points, and all three Holm-adjusted p-values are 1.0000. The most informative quantity in this table is the ratio of seed dispersions, not any single contrast: the seed standard deviation of Qwen is 6.2 points against 0.5 points for ResNet-50, so the two families differ more in stability than in central tendency.

\begin{table}[htbp]
\centering
\setlength{\tabcolsep}{4pt}
\renewcommand{\arraystretch}{1.15}
\caption*{Table 28. Tower14 Swin-T control (seed 42; n = 5,536; 14 test towers). Differences are Qwen minus Swin-T. This is a supplementary single-seed control rather than a three-seed replication.}
\begin{tabularx}{\textwidth}{>{\hsize=1.048\hsize\RaggedRight\hyphenpenalty=10000\exhyphenpenalty=50\arraybackslash}X >{\hsize=1.024\hsize\RaggedRight\hyphenpenalty=10000\exhyphenpenalty=50\arraybackslash}X >{\hsize=1.059\hsize\RaggedRight\hyphenpenalty=10000\exhyphenpenalty=50\arraybackslash}X >{\hsize=0.927\hsize\RaggedRight\hyphenpenalty=10000\exhyphenpenalty=50\arraybackslash}X >{\hsize=1.354\hsize\RaggedRight\hyphenpenalty=10000\exhyphenpenalty=50\arraybackslash}X >{\hsize=1.030\hsize\RaggedRight\hyphenpenalty=10000\exhyphenpenalty=50\arraybackslash}X >{\hsize=0.558\hsize\RaggedRight\hyphenpenalty=10000\exhyphenpenalty=50\arraybackslash}X}
\toprule
Label space & Qwen3-VL-8B & ResNet-50 448 px & Swin-T 448 px & Qwen–Swin difference & 95\% CI & Raw p \\
\midrule
Common six & 0.9369 & 0.9561 & 0.9695 & −3.26 pt & [−6.42, −0.11] & 0.0191 \\
Full seven & 0.9459 & 0.8195 & 0.9024 & +4.35 pt & [−4.73, +13.42] & 0.9828 \\
\bottomrule
\end{tabularx}
\end{table}

The single-seed Swin-T control reaches 0.9695 common-six macro recall, above both the Qwen and the ResNet-50 seed-42 runs, and exceeds ResNet-50 by 1.34 points. Under the full seven-class label space Qwen is above Swin-T and the interval includes zero. The tower ordering is therefore sensitive to backbone, seed, and label space and identifies no stable winner. Because only one Swin-T seed is available, this control closes a baseline-coverage gap rather than adding a replicated family-level claim, and the three-split result of Table 23 is unchanged. Figure 8 shows the tower14 results across seeds and label spaces.

\begin{figure}[!htbp]
\centering
\IfFileExists{figures/figure8.jpg}{\includegraphics[width=0.95\linewidth]{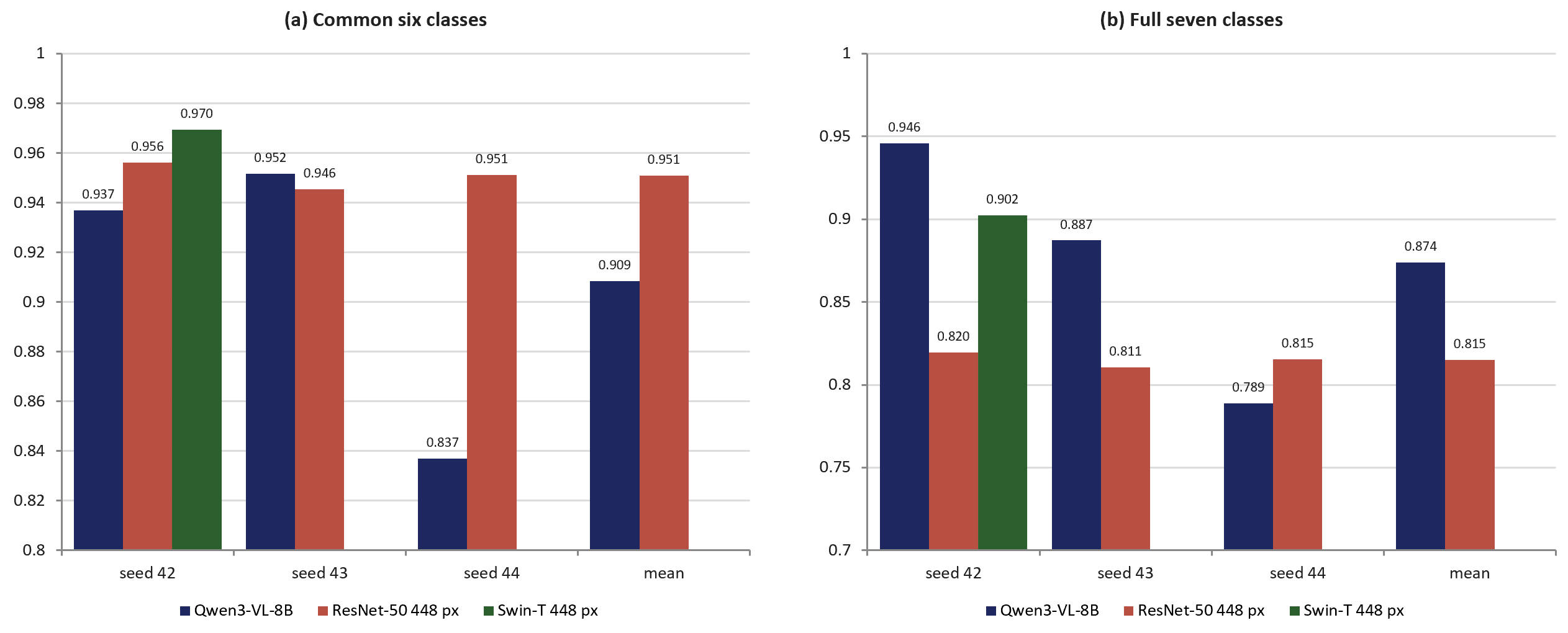}}{\fbox{\parbox{0.92\linewidth}{\centering\vspace{2.5cm}\small Placeholder for \texttt{figures/figure8.jpg}\vspace{2.5cm}}}}
\caption*{Figure 8. Tower14 replication and Swin-T control. Seed-wise common-six and full-seven macro recall for Qwen3-VL-8B, ResNet-50, and Swin-T, from Tables 27 and 28, with each family's seed standard deviation annotated.}
\end{figure}
\FloatBarrier

Diagnosis of the seed-44 result. The drop at seed 44 is not a training or inference failure. Final training loss is 0.01711 and token accuracy approximately 0.998, close to the other seeds, and a rerun on a healthy accelerator reproduces the same two \texttt{torned-up} errors while changing common-six macro recall by 0.08 points. The binary prompt still classifies both images as defective; the error occurs only when choosing between \texttt{good} and \texttt{torned-up} in the multiclass prompt. Excluding the three-item \texttt{torned-up} class leaves seed 44 at 0.9377 five-class macro recall against 0.9243 and 0.9420 for seeds 42 and 43. The result is therefore rare-class seed sensitivity amplified by macro averaging rather than general model degradation, which is the same mechanism identified in Section 5.7.2, observed here in a replication design.

\section*{6. Discussion}
\subsection*{6.1. What the Controlled Comparisons Establish}
The audit produces one negative result at binary screening, one artefact identification at seven-way typing, and one difference that the controls bound without explaining.

At binary screening on a matched partition at full training data, the best vision-only backbone and the best adapted VLM differ by 0.03 points, and McNemar's test returns p = 1.000 on 48 discordant items split 24 to 24. This is a balanced discordant-pair result rather than a formal equivalence test, and it supports the removal of the 13-to-15-point lead seen in the uncontrolled comparison rather than the equivalence of the two families.

Most of the seven-way gap against InternVL3.5-8B is attributable to input pixel budget. Within a single backbone, moving from 224 px to 672 px changes the gap from +20.53 to −0.57 points for ResNet-50 and from +23.67 to +4.70 points for Swin-T, and the measured matched budget of approximately 768 px places InternVL between the two ResNet-50 values. Because the vision-only sweep was evaluated on the test set, the 672 px and 896 px settings rather than the highest-scoring 448 px setting are the controlled comparison.

Nothing in the split analysis supports a family-level ordering. After split-specific retraining, Qwen is 0.51 points behind at crop level and 4.40 points behind at image level. The nine-tower split retains a +8.02-point advantage whose Holm-adjusted p in the three-split family is 0.0717. The 14-tower, three-seed replication reverses the picture: ResNet-50 leads in two of three matched seeds, and the two families differ more in seed stability, 0.5 against 6.2 points of standard deviation, than in mean. On the original T3 set the Qwen–ResNet-50 448 px effect is +7.99 points with a tower-clustered Holm p of 0.834, and nine test items carry +7.85 points of it. Only the contrast against Swin-T at 672 px survives correction anywhere in that family.

One difference remains bounded but unexplained. Qwen3-VL-8B exceeds InternVL3.5-8B by 11.18 points at seven-way typing and 1.3 points at binary screening under identical training and evaluation. Source-pixel matching does not remove it, and a visual-token account runs the wrong way, since Qwen leads while consuming 56\% fewer visual tokens. Resolution, language grounding, and scale do not account for this difference on the evidence collected here, which is a weaker statement than excluding them as contributors.

The methodological point is straightforward. A comparison between a language-grounded and a vision-only model is not interpretable unless it reports its evaluation partition, evaluated item set, label space, replication count, input resolution, and side information. Even a comparison controlled on all six axes can leave a model-specific difference that none of them explains. Five of the six axes were mismatched at some point in the successive passes of this study. The sixth, side information, cannot be matched between the two families, so we declare it rather than control it.

\subsection*{6.2. Why the Seven-Way Result Is Not an Architecture Result}
Several candidate explanations were tested individually. Input pixel budget accounts for most of the gap against the InternVL models, both within a single vision backbone and on the VLM side, where InternVL itself varies by 6.16 points across source-pixel budgets. Language grounding is necessary for both adapted VLMs, since each collapses to 14.29\% without the decision tree, but it is shared between them and therefore cannot distinguish them. Model scale within a family contributes at most 3.74 points from 2B to 8B. Swin-T being a transformer rules out the coarse convolutional-versus-transformer contrast but does not exclude architecture-specific differences in visual encoding.

What these controls fail to explain is the difference between Qwen3-VL-8B and InternVL3.5-8B. The pixel-budget audit establishes that Qwen is sensitive to its input budget and that the two models remain separated at every matched source-pixel setting, and the token statistics logged in the same audit exclude the most obvious remaining account: at native settings the leading model uses 329.5 mean visual tokens against 752.9 for the trailing model. An account in which the stronger model simply resolves more of the image is inconsistent with that measurement. Tokenizer design, patch-formation policy, pre-training composition, and architecture-specific encoding remain open, and a matched-visual-token comparison would separate them.

A separate point concerns the metric. Two classes holding nine test items between them move the seven-class macro-average by 22.9 points in opposite directions in one configuration and account for +7.85 of a +7.99-point gap in another. The mechanism replicates across configurations and across the split analysis, where \texttt{nest} and \texttt{missing-cap} rather than a single four-item class carry the tower-level difference. Unweighted macro-averaging over a long-tailed label set converts rare-class sampling into headline differences, and which rare class does the converting changes with the split.

\subsection*{6.3. Practical Implications}
For binary screening under a fixed, well-populated label set, a fine-tuned vision backbone matches the best adapted VLM while providing substantially higher measured batch throughput, so on this evidence there is no accuracy argument for the VLM. For fine-grained typing the picture is split- and seed-dependent rather than family-dependent. Against the InternVL models, parity is restored by raising the pixel budget of the vision backbone. Against Qwen3-VL-8B, the retrained crop and image comparisons favour ResNet-50 by 0.51 and 4.40 points with neither difference significant, the nine-tower comparison favours Qwen but does not survive correction, and the 14-tower replication favours ResNet-50 in two of three seeds. A deployment decision should therefore treat model choice at this granularity as unresolved.

Two operational conclusions do not depend on resolving it. First, seed stability is a selection criterion in its own right. Across the tower14 replication the vision-only backbone varies by 0.5 points between seeds and the VLM by 6.2 points, and a single rare-class error moves a seven-class macro-average by more than ten points, so a conclusion drawn from one training run may be reversed by a second. Second, the distinctive property of the VLM is the side-information channel quantified by the prompt ablation: restoring the decision tree at inference time, with no retraining, recovers 54.97 to 70.07 points depending on the model. An operator who needs to add a defect category or revise its criteria can do so by editing text. That flexibility, rather than a measured accuracy advantage, is what the evidence here supports for the VLM family.

The cost analysis carries its own caution. Balanced accuracy cannot separate error profiles dominated by misses, hallucinations, or abstentions, and all three cases are occupied by real models in this study, including one that leaves 20\% of defective items unjudged while reporting a zero miss rate. Where such a model is selected as risk-minimizing, the selection is conditional on pricing manual review at zero and on the constructed 37.9\% defect prevalence of the benchmark. Priced and re-based realistically, the crossover moves beyond any operating point we would defend as routine. The practical recommendation follows Provost and Fawcett \cite{ref11}: report the sensitivity surface rather than the operating point.

\subsection*{6.4. Limitations}
The principal limitation concerns data validity. The benchmark groups records by a filename-derived \texttt{parent\_id}, but 97.7\% of T2 test items and 99.6\% of T3 test items share a source-photograph key with at least one training item, so \texttt{parent\_id} disjointness prevents exact parent-group overlap without guaranteeing global source-photograph disjointness. In the crop and image diagnostics of Section 5.9 the situation was initially more severe, because the reused checkpoints had previously seen about 70\% of the items in those test files; those rows have been replaced by retrained ones. The tower-level models were trained on tower-disjoint splits and are the only rows used as split-generalization evidence. Three further data-side constraints apply: a subset of the supervised data spans October 2020 to June 2021 with non-uniform defect composition, annotation noise is unquantified, and T1 is constructed but not evaluated, so the scope of the study is T2 and T3 on a single dataset with no cross-domain evaluation.

The second group of limitations concerns statistical strength and experimental coverage. The primary Qwen3-VL-8B and InternVL3.5-8B global results have two seeds each, and the 2B adapted arm remains single-seed. An earlier constant-output result was traced to an evaluation-input construction error that omitted image content rather than to a training failure; after re-evaluation with image-bearing messages, InternVL seed 41 reaches 0.9586 T2 balanced accuracy and 0.7440 seven-way macro recall. Two seeds do not support a precise seed-variance estimate, and the tower14 replication shows why that matters, since the seed standard deviation of the VLM there is 6.2 points. Only Qwen3-VL-8B and ResNet-50 have three matched tower14 seeds, and the Swin-T control is single-seed. The crop-level regime still allows same-image and same-tower overlap and the image-level regime retains tower overlap, so neither is an unseen-tower test. The tower14 replication remains exploratory rather than pre-registered. The source-pixel-matched InternVL control is not a visual-token match. The 448 px vision-only configuration was selected on the test set, and the measured matched budget of approximately 768 px falls between rather than on the available settings. The multi-task and epoch ablations use a 200-item subsample whose T3 diagnostic covers five of the seven classes. The two API-served models are hosted services without version pinning, as recorded in Table A5b. Finally, the instrumented pixel-budget pass differs from the main adaptation pass by 0.31 and 0.56 points on seven-way macro recall for the two 8B models, and every comparison is made within a single pass.

Deployment coverage is the third limitation. Vision-only memory, single-image latency, and sustained batch throughput were measured, but energy consumption, long-duration thermal stability, and performance on UAV-representative accelerators were not, and all inference ran on a workstation-class GPU with preloaded tensors. The VLM and vision-only throughput figures were obtained with different harnesses and are not directly comparable. All models operate on ground-truth crops rather than full UAV frames, so this is a classification audit rather than an end-to-end inspection system. The UAV-specific elements are the data source and the deployment constraints, while the methodology applies to industrial defect classification more broadly.

\subsection*{6.5. Future Work}
Four extensions would most directly strengthen the unresolved claims. First, Qwen and InternVL should be compared at matched visual-token budgets as well as matched source-pixel budgets, because the token statistics in Table 16 show the difference running opposite to a token-budget account and the mechanism cannot be isolated without that control. Second, tower-level replication should be pre-registered with more lines and towers, since the tower14 result shows that a new split can reverse the sign of the earlier tower advantage. Third, all seven-way metrics should be regenerated in a single instrumented pass so that balanced accuracy, macro-F1, and per-class recall for every model come from one evaluation. Fourth, deployment conclusions require energy, thermal-stability, and embedded-accelerator measurements under a common harness for both families.

Two longer-horizon directions follow from the framework itself. Extracting a calibrated confidence signal from each VLM would permit full cost-curve comparison rather than the single-operating-point analysis used here. Replacing ground-truth boxes with a trained detector would close the loop on realistic deployment while making use of the released T1 split.

\section*{7. Conclusions}
This study compared vision–language and vision-only models for UAV power-line defect assessment and found that most headline differences are properties of the evaluation protocol rather than of the models.

At binary screening, mismatched test files produced a VLM lead of 13 to 15 points, whereas the identical 2,021-item partition produced a tie of 97.17\% against 97.14\% with McNemar p = 1.000. At seven-way typing, moving ResNet-50 and Swin-T from 224 px to the settings nearest the measured pixel budget changes the gap against InternVL3.5-8B from +20.53 to −0.57 points and from +23.67 to +4.70 points respectively, placing InternVL between the two ResNet-50 values.

A pixel-budget audit that logs grids, token counts, and input hashes moves Qwen3-VL-8B macro recall from 78.61\% at a 65,536-pixel budget to 89.39\% at native resolution. A source-pixel-matched InternVL control varies by 6.16 points and remains 7.43 to 13.61 points below Qwen, and the token statistics from the same audit show Qwen leading while consuming 56\% fewer visual tokens. Neither source pixels nor token budget accounts for the difference between the two VLMs, which the controls bound rather than explain.

After split-specific retraining, common-class alignment, cluster-aware randomization inference, and Holm correction, no split regime provides a significant Qwen–ResNet advantage. Crop and image comparisons favour ResNet-50. The nine-tower split favours Qwen by 8.02 points at seed 42 without surviving correction, and the 14-tower three-seed replication favours ResNet-50 in two of three seeds, with Qwen averaging 0.9085 against 0.9509 for ResNet-50 on the common six classes. The two families differ more in seed stability, 6.2 against 0.5 points of standard deviation, than in central tendency. A single-seed Swin-T control reaches 0.9695 on the common-six tower partition but 0.9024 under the full seven-class label space, so the tower ordering is label-space-dependent as well.

Language grounding is necessary, but it does not distinguish the two VLMs. Without the defect decision tree, both adapted VLMs collapse to an identical 14.29\%, the value expected from majority-class prediction. Balanced accuracy also hides materially different miss, hallucination, and abstention regimes: a cost analysis that prices abstention at zero selects a model that leaves 20\% of defective items unjudged, and with realistic review costs and field prevalence that selection no longer holds.

The broader contribution is an audited benchmark protocol, not a family-level winner. Partition, evaluated item set, label space, seed, split construction, input resolution, and side information must be reported together, or a comparison between a VLM and a vision-only model is not interpretable. Two findings survive the audit and are worth carrying forward independently of it: an unweighted macro-average over a label set whose rarest classes hold single-digit test support can be moved more than ten points by a handful of items, and the run-to-run instability of an adapted VLM can exceed the family-level difference it is being used to measure. The difference between the two adapted VLMs is real in magnitude and unresolved in mechanism, and future work should test it at matched visual-token budgets.

\section*{Author Contributions}
Conceptualization, L.Z. and S.X.; methodology, L.Z., S.X. and Y.Y.; software, L.Z. and J.K.; validation, J.K. and Y.Y.; formal analysis, L.Z. and Y.Y.; investigation, L.Z., J.K. and Y.Y.; resources, S.X.; data curation, J.K. and Y.Y.; writing—original draft preparation, L.Z.; writing—review and editing, L.Z., S.X., J.K. and Y.Y.; visualization, J.K.; supervision, S.X. and Y.Y.; project administration, S.X.; funding acquisition, S.X. All authors have read and agreed to the published version of the manuscript.

The division of work follows the structure of the audit. L.Z. designed the six-axis parity protocol and the cost-sensitive risk formulation with abstention, implemented the evaluation harness and the adaptation pipeline for both the VLM and the vision-only arms, and ran the resolution control, the instrumented pixel-budget pass, the prompt ablation, and the split-level retraining experiments. S.X. conceived and coordinated the study, provided the data and computational resources, and supervised the audit protocol, the statistical design, and the multiple-comparison correction. J.K. built the ElecVQA-Bench construction pipeline from the InsPLAD release, curated the split manifests with their SHA-256 digests and the per-class support tables, instrumented the grid, token-count and input-hash logging, and prepared the tables and figures. Y.Y. developed the prevalence-sensitivity analysis and the cluster-aware randomization inference, performed the quantization and deployment-envelope measurements, and independently reproduced the tower-level three-seed replication. All authors contributed to the interpretation of the results and to revising the manuscript.

\section*{Funding}
This research was funded by the Science and Technology Project of State Grid Sichuan Electric Power Company, grant number 521997240003.

\section*{Institutional Review Board Statement}
Not applicable.

\section*{Informed Consent Statement}
Not applicable.

\section*{Data Availability Statement}
Construction scripts, split manifests with SHA-256 digests, prompt templates, the evaluation harness, per-class support tables, the construction audit log, the prediction files carrying the hashes reported in Tables 15 and 16, and LoRA adapter weights will be released in a public repository and archived with a DOI upon acceptance. Source imagery is the public InsPLAD dataset \cite{ref1}, is used under its public release terms, and is not redistributed. Derived metadata, code, and adapter weights will be released under licences compatible with those terms.

\section*{Acknowledgments}
The authors thank the maintainers of the InsPLAD dataset for releasing the source imagery and annotations.

\section*{Conflicts of Interest}
The authors declare no conflicts of interest.

\section*{Abbreviations}
\begin{table}[htbp]
\centering
\setlength{\tabcolsep}{3.0pt}
\renewcommand{\arraystretch}{1.15}
\begin{tabularx}{0.567\textwidth}{>{\hsize=0.519\hsize\RaggedRight\hyphenpenalty=10000\exhyphenpenalty=50\arraybackslash}X >{\hsize=1.481\hsize\RaggedRight\hyphenpenalty=10000\exhyphenpenalty=50\arraybackslash}X}
\toprule
Abbreviation & Meaning \\
\midrule
BalAcc & Balanced accuracy \\
bf16 & Brain floating-point, 16-bit \\
CI & Confidence interval \\
CNN & Convolutional neural network \\
EAR & Error-asymmetry ratio \\
FP8 & Floating-point, 8-bit \\
INT8 & Integer, 8-bit \\
LoRA & Low-rank adaptation \\
NF4 & 4-bit NormalFloat \\
QLoRA & Quantized low-rank adaptation \\
ROI & Region of interest \\
S200 & Shared 200-item evaluation subsample \\
SD & Standard deviation \\
TNR & True negative rate \\
TPR & True positive rate \\
UAV & Unmanned aerial vehicle \\
ViT & Vision transformer \\
VLM & Vision–language model \\
VQA & Visual question answering \\
\bottomrule
\end{tabularx}
\end{table}

\section*{Appendix A}
\begin{table}[htbp]
\centering
\setlength{\tabcolsep}{3.0pt}
\renewcommand{\arraystretch}{1.15}
\caption*{Table A1. Notation.}
\begin{tabularx}{0.619\textwidth}{>{\hsize=0.614\hsize\RaggedRight\hyphenpenalty=10000\exhyphenpenalty=50\arraybackslash}X >{\hsize=1.386\hsize\RaggedRight\hyphenpenalty=10000\exhyphenpenalty=50\arraybackslash}X}
\toprule
Symbol & Meaning \\
\midrule
Y ∈ \{N, D\} & ground truth, normal or defective \\
Ŷ ∈ \{N, D, A\} & prediction, normal, defective, or abstain \\
α, β & hallucination rate, miss rate \\
γ\textsubscript{D}, γ\textsubscript{N} & abstention rates by true class \\
π\textsubscript{D}, π\textsubscript{N} & class priors of the evaluation set \\
C\textsubscript{m}, C\textsubscript{h}, C\textsubscript{a} & cost of a miss, a false alarm, an abstention \\
ρ = C\textsubscript{m}/C\textsubscript{h} & miss-to-false-alarm cost ratio \\
κ = C\textsubscript{a}/C\textsubscript{h} & abstention cost ratio \\
R\textsubscript{i}(ρ, κ) & expected risk of model i, Equation (2) \\
ρ*(κ) & crossover cost ratio, Equation (3) \\
K & number of ground-truth answer classes \\
Δ\textsubscript{b} & set of error profiles consistent with BalAcc = b \\
S200 & shared 200-item evaluation subsample \\
\bottomrule
\end{tabularx}
\end{table}

\begingroup
\setlength{\tabcolsep}{4pt}
\renewcommand{\arraystretch}{1.15}
\begin{xltabular}{\textwidth}{>{\hsize=0.625\hsize\RaggedRight\hyphenpenalty=10000\exhyphenpenalty=50\arraybackslash}X >{\hsize=1.030\hsize\RaggedRight\hyphenpenalty=10000\exhyphenpenalty=50\arraybackslash}X >{\hsize=1.048\hsize\RaggedRight\hyphenpenalty=10000\exhyphenpenalty=50\arraybackslash}X >{\hsize=0.703\hsize\RaggedRight\hyphenpenalty=10000\exhyphenpenalty=50\arraybackslash}X >{\hsize=1.188\hsize\RaggedRight\hyphenpenalty=10000\exhyphenpenalty=50\arraybackslash}X >{\hsize=0.784\hsize\RaggedRight\hyphenpenalty=10000\exhyphenpenalty=50\arraybackslash}X >{\hsize=1.283\hsize\RaggedRight\hyphenpenalty=10000\exhyphenpenalty=50\arraybackslash}X >{\hsize=1.339\hsize\RaggedRight\hyphenpenalty=10000\exhyphenpenalty=50\arraybackslash}X}
\caption*{Table A2. Experiment, configuration, and evaluation crosswalk. Where this table disagrees with a caption, this table governs. The pass column distinguishes the main adaptation pass from the instrumented pixel-budget pass described in Section 4.6.}\\
\toprule
Table & Models & Evaluated on & n & Class set & Seeds & Pass & Comparable to \\
\midrule
\endfirsthead
\toprule
Table & Models & Evaluated on & n & Class set & Seeds & Pass & Comparable to \\
\midrule
\endhead
2, 3 & 7 VLMs, zero-shot & T2 test & 2,021 & binary & 1 & main & Table 4 \\
4, 6 & 3 VLMs + LoRA & T2 / T3 test & 2,021 / 7,284 & binary / K = 7 & 1 & main & Tables 2, 9, 10 \\
5 & Qwen, InternVL + LoRA & T2 / T3 test & 2,021 / 7,284 & binary / K = 7 & 2 each & main & Table 4 \\
7, A8 & InternVL-2B + LoRA & S200 & 200 & binary / five-class T3 & 1 & main & within-table only \\
8 & ResNet-50, Swin-T & T2 test & 2,021 & binary & 3 & — & within-table only \\
9 & both families & T2 test & 2,021 & binary & 3 / 1 & main & Table 4, Table 10 row K = 2 \\
10, K = 2 & both families & T2 test & 2,021 & binary & 3 / 1 & main & Table 9 \\
10, K = 3 & both families & T3 test & 7,284 & option-level, not a collapse of K = 7 & 1 & main & within-row only \\
10, K = 7 & both families & T3 test & 7,284 & K = 7 & 1 & main & Tables 11, 17, 18 \\
11, 12 & ResNet-50, Swin-T, VLMs & T3 test & 7,284 & K = 7, raw-accuracy McNemar & 1 & main & Table 10 row K = 7 \\
13 & Qwen vs. vision-only & T3 test & 7,284 & K = 7, tower-clustered macro recall & 1 & main & Tables 11, 14 \\
14 & Qwen3-VL-8B & T3 test & 7,284 & K = 7 per-class recall & 1 & main & Table 13 \\
15 & Qwen3-VL-8B & T3 test & 7,284 & K = 7, audited source-pixel budgets & 1 & instrumented & Table 16 \\
16 & Qwen, InternVL & T3 test & 7,284 & K = 7, source-pixel-matched & 1 & instrumented & Table 15 \\
17 & InternVL, Swin-T 448 px & T3 test & 7,284 & K = 7 per-class recall & 1 & main & Table 14, different configuration \\
18, 19 & InternVL, Qwen + LoRA & T3 test & 7,284 & K = 7, prompt arms & 1 & main & Table 10 VLM columns \\
21 & InternVL-8B + LoRA & T2 / T3 test & 2,021 / 7,284 & binary / K = 7 & 1 & main & Table 4 \\
22 & ResNet-50, Swin-T & 512 sampled images & 512 & deployment only & 1 & — & within-table only \\
23, 24, 25 & Qwen, ResNet-50 & crop / image / tower T3 & 6,672 / 6,620 / 13,574 & common six & 1 / 1 / see Table 26 & retrained & within-split only \\
26 & Qwen, ResNet-50 & tower T3 & 13,574 & common six, seed sensitivity & Qwen 2; ResNet 3 & retrained & Table 23 tower row \\
27, 28 & Qwen, ResNet-50, Swin-T & tower14 T3-only & 5,536 & common six primary; full seven secondary & 3 / 3 / 1 & retrained & within-table only \\
\bottomrule
\end{xltabular}
\endgroup

S200 is a fixed 200-item subsample drawn once with seed 42 and used only in Tables 7 and A8. The value 0.949 for InternVL-2B in Table 4 and 0.932 in Table 7 differ because the first is the full 2,021-item test set and the second is S200. The crop, image, tower, and tower14 test sets are separate split diagnostics and are not comparable to one another or to the original T3 test set.

\begin{table}[htbp]
\centering
\setlength{\tabcolsep}{4pt}
\renewcommand{\arraystretch}{1.15}
\caption*{Table A3. Label crosswalk between tasks.}
\begin{tabularx}{\textwidth}{>{\hsize=0.980\hsize\RaggedRight\hyphenpenalty=10000\exhyphenpenalty=50\arraybackslash}X >{\hsize=0.766\hsize\RaggedRight\hyphenpenalty=10000\exhyphenpenalty=50\arraybackslash}X >{\hsize=0.766\hsize\RaggedRight\hyphenpenalty=10000\exhyphenpenalty=50\arraybackslash}X >{\hsize=1.488\hsize\RaggedRight\hyphenpenalty=10000\exhyphenpenalty=50\arraybackslash}X}
\toprule
InsPLAD source label & T2 label (K = 2) & T3 label (K = 7) & Definition \\
\midrule
good & normal & good & No visible defect \\
rust & defect & rust & Localized oxidation on metal fittings \\
corrosão & defect & corrosão & Generalized corrosion of metallic surfaces \\
peeling-paint & defect & peeling-paint & Loss of protective coating \\
missing-cap & defect & missing-cap & Missing or displaced component cap \\
torned-up & defect & torned-up & Torn or severed element \\
nest & defect & nest & Foreign object, bird nest, on the asset \\
\bottomrule
\end{tabularx}
\end{table}

\begin{table}[htbp]
\centering
\setlength{\tabcolsep}{3.0pt}
\renewcommand{\arraystretch}{1.15}
\caption*{Table A4. Per-class support, original test partitions.}
\begin{tabularx}{0.357\textwidth}{>{\hsize=1.097\hsize\RaggedRight\hyphenpenalty=10000\exhyphenpenalty=50\arraybackslash}X >{\hsize=1.339\hsize\RaggedRight\hyphenpenalty=10000\exhyphenpenalty=50\arraybackslash}X >{\hsize=0.564\hsize\RaggedRight\hyphenpenalty=10000\exhyphenpenalty=50\arraybackslash}X}
\toprule
Partition & Class & n \\
\midrule
T2 (2,021) & normal & 1,255 \\
T2 (2,021) & defect & 766 \\
T3 (7,284) & good & 6,518 \\
T3 (7,284) & rust & 334 \\
T3 (7,284) & corrosão & 184 \\
T3 (7,284) & missing-cap & 136 \\
T3 (7,284) & nest & 103 \\
T3 (7,284) & torned-up & 5 \\
T3 (7,284) & peeling-paint & 4 \\
\bottomrule
\end{tabularx}
\end{table}

\begin{table}[htbp]
\centering
\setlength{\tabcolsep}{3.0pt}
\renewcommand{\arraystretch}{1.15}
\caption*{Table A5. Software and run configuration.}
\begin{tabularx}{0.654\textwidth}{>{\hsize=0.691\hsize\RaggedRight\hyphenpenalty=10000\exhyphenpenalty=50\arraybackslash}X >{\hsize=1.309\hsize\RaggedRight\hyphenpenalty=10000\exhyphenpenalty=50\arraybackslash}X}
\toprule
Component & Setting \\
\midrule
Software & PyTorch 2.11.0+cu128; Transformers 5.14.1; PEFT 0.20.0; Swift 4.5.2; bitsandbytes weight-only quantization \\
Accelerator & 1 × NVIDIA RTX PRO 6000 Blackwell (96 GB) \\
Training precision & bf16 \\
LoRA & rank 32, α = 64, dropout 0.05; target modules q\_proj, k\_proj, v\_proj, o\_proj, gate\_proj, up\_proj, down\_proj; vision encoder frozen \\
Optimizer & AdamW; VLM learning rate 1 × 10⁻⁴, cosine schedule, 3\% warmup, effective batch 16, weight decay 0.01 \\
Checkpoint selection & final checkpoint after the fixed epoch budget; no early stopping \\
Loss & VLM autoregressive token loss; vision-only class-weighted cross-entropy \\
Epochs & 3 for VLMs; 10 for vision-only baselines \\
Vision-only input & 224 / 448 / 672 / 896 px. Training: resize, random horizontal flip, random rotation 10°, colour jitter 0.2/0.2/0.2, ImageNet normalization. Evaluation: resize and ImageNet normalization. \\
VLM input & 448 px tiles, dynamic tiling to 12 tiles, 256 tokens per tile. Audited native means on T3: InternVL3.5-8B 2.94 patches and 752.9 visual tokens; Qwen3-VL-8B 329.5 mean and 144 median visual tokens. \\
Hosted API models & see Table A5b; endpoint, model identifier, and access date recorded; hosted weights not version-pinned \\
Quantization & bitsandbytes NF4 and INT8 weight-only, bf16 activations \\
Seeds & global VLM 41 and 42; vision-only training-budget sweep 3 seeds; tower14 42, 43, 44 \\
\bottomrule
\end{tabularx}
\end{table}

\begin{table}[htbp]
\centering
\setlength{\tabcolsep}{4pt}
\renewcommand{\arraystretch}{1.15}
\caption*{Table A5b. Hosted API provenance.}
\begin{tabularx}{\textwidth}{>{\hsize=0.486\hsize\RaggedRight\hyphenpenalty=10000\exhyphenpenalty=50\arraybackslash}X >{\hsize=1.681\hsize\RaggedRight\hyphenpenalty=10000\exhyphenpenalty=50\arraybackslash}X >{\hsize=0.362\hsize\RaggedRight\hyphenpenalty=10000\exhyphenpenalty=50\arraybackslash}X >{\hsize=1.471\hsize\RaggedRight\hyphenpenalty=10000\exhyphenpenalty=50\arraybackslash}X}
\toprule
Model identifier & Endpoint & Access date & Reproducibility status \\
\midrule
Qwen3.8-27B & \texttt{<internal-\allowbreak vllm-\allowbreak endpoint-\allowbreak redacted>} & 2026-08-24 & service-reported model root \texttt{/root/data-tmp/Qwen3.8-27B}; no weight-revision hash \\
Qwen3.5-122B-A10B-FP8 & \texttt{<internal-\allowbreak vllm-\allowbreak endpoint-\allowbreak redacted>} & 2026-08-24 & service identifier only; no weight-revision hash \\
\bottomrule
\end{tabularx}
\end{table}

The hosted services do not expose a stable weight-revision hash, so the corresponding zero-shot results are provenance snapshots rather than version-pinned reproductions.

\begin{table}[htbp]
\centering
\setlength{\tabcolsep}{4pt}
\renewcommand{\arraystretch}{1.05}
\caption*{Table A6. Experiment ledger. Rows marked as included are accounted for within another row's budget.}
\begin{tabularx}{\textwidth}{>{\hsize=0.790\hsize\RaggedRight\hyphenpenalty=10000\exhyphenpenalty=50\arraybackslash}X >{\hsize=1.535\hsize\RaggedRight\hyphenpenalty=10000\exhyphenpenalty=50\arraybackslash}X >{\hsize=0.503\hsize\RaggedRight\hyphenpenalty=10000\exhyphenpenalty=50\arraybackslash}X >{\hsize=1.190\hsize\RaggedRight\hyphenpenalty=10000\exhyphenpenalty=50\arraybackslash}X >{\hsize=0.982\hsize\RaggedRight\hyphenpenalty=10000\exhyphenpenalty=50\arraybackslash}X}
\toprule
Group & Experiment & GPU-h & Evaluated set & Seeds \\
\midrule
Construction & Audit, split, ROI and VQA construction & 0 & — & — \\
Zero-shot & 7-model zero-shot & 0 & T2 2,021; T3 7,284 & 1 \\
Adaptation & LoRA adaptation, 8B, 2B, Qwen & 41 & T2 2,021; T3 7,284; S200 & 1 \\
Baseline & Vision-only baseline, superseded & 2 & 7,851 & 1 \\
Budget sweep & Vision-only training budget, re-scored & 6 & T2 2,021 & 3 \\
T3 baseline & Vision-only T3 baseline & 4 & T3 7,284 & 1 \\
Ablation & Corpus and epoch ablations & 38 & S200 & 1 \\
Quantization & INT8 and NF4, full test & 4 & T2 2,021; T3 7,284 & 1 \\
Resolution & Resolution sweep, 224/448/672/896 px & 2 & T3 7,284 & 1 \\
Tile control & InternVL single-tile control & 1 & T3 7,284 & 1 \\
Pixel budget & Qwen pixel-budget audit & 1.8 & T3 7,284 & 1 \\
Pixel budget & InternVL source-pixel control & 2.0 & T3 7,284 & 1 \\
Prompt & Prompt ablation, InternVL, three arms & 10 & T3 7,284 & 1 \\
Prompt & Prompt ablation, Qwen replication & 1 & T3 7,284 & 1 \\
Inference & McNemar and cluster-aware randomization & 0 (CPU) & existing predictions & — \\
Verification & Independent-path verification and leakage audit & 0 & T2 200 / full & — \\
Split & Crop and image split-specific retraining & included & crop and image T3 tests & 1 \\
Split & Tower split construction and retraining & 7 & tower T3 13,574 & Qwen 2; ResNet 3 \\
Replication & Tower14 three-seed replication & 65 & tower14 T3-only 5,536 & 3 / 3 \\
Replication & Tower14 Swin-T control & 0.5 & tower14 T3-only 5,536 & 1 \\
Deployment & Vision-only deployment benchmark & 0.2 & 512 images at 448/672 px & 1 \\
Replication & Qwen global seed-41 replication & 11 & T2 2,021; T3 7,284 & 2 \\
Total &  & ≈205 &  &  \\
\bottomrule
\end{tabularx}
\end{table}

\begin{table}[htbp]
\centering
\setlength{\tabcolsep}{3.0pt}
\renewcommand{\arraystretch}{1.15}
\caption*{Table A7. Sensitivity of the error-asymmetry ratio to the smoothing constant.}
\begin{tabularx}{0.502\textwidth}{>{\hsize=1.544\hsize\RaggedRight\hyphenpenalty=10000\exhyphenpenalty=50\arraybackslash}X >{\hsize=0.711\hsize\RaggedRight\hyphenpenalty=10000\exhyphenpenalty=50\arraybackslash}X >{\hsize=0.819\hsize\RaggedRight\hyphenpenalty=10000\exhyphenpenalty=50\arraybackslash}X >{\hsize=0.926\hsize\RaggedRight\hyphenpenalty=10000\exhyphenpenalty=50\arraybackslash}X}
\toprule
Model & ε = 0.1 & ε = 0.01 & ε = 0.001 \\
\midrule
InternVL3.5-2B & 0.12 & 0.01 & 0.00 \\
Qwen3-VL-2B & 0.16 & 0.03 & 0.02 \\
Qwen2.5-VL-7B & 0.45 & 0.30 & 0.28 \\
Qwen3.8-27B & 6.80 & 33.41 & 60.93 \\
InternVL3.5-8B & 6.94 & 41.23 & 96.09 \\
Qwen3.5-122B & 3.89 & 8.96 & 10.65 \\
Qwen3-VL-8B & 4.13 & 13.30 & 18.40 \\
\bottomrule
\end{tabularx}
\end{table}

The value for InternVL3.5-8B varies by a factor of fourteen across this range and the ordering of the 27B and 122B models is not preserved.

\begin{table}[htbp]
\centering
\setlength{\tabcolsep}{3.0pt}
\renewcommand{\arraystretch}{1.15}
\caption*{Table A8. S200 five-class T3 diagnostic, InternVL3.5-2B + LoRA. S200 contains 177 \texttt{good}, 12 \texttt{rust}, 7 \texttt{missing-cap}, 3 \texttt{corrosão}, and 1 \texttt{nest} item, and no instances of \texttt{peeling-paint} or \texttt{torned-up}. These values are therefore a five-class diagnostic and are not comparable to any seven-class result in this paper.}
\begin{tabularx}{0.772\textwidth}{>{\hsize=1.054\hsize\RaggedRight\hyphenpenalty=10000\exhyphenpenalty=50\arraybackslash}X >{\hsize=0.886\hsize\RaggedRight\hyphenpenalty=10000\exhyphenpenalty=50\arraybackslash}X >{\hsize=0.574\hsize\RaggedRight\hyphenpenalty=10000\exhyphenpenalty=50\arraybackslash}X >{\hsize=1.805\hsize\RaggedRight\hyphenpenalty=10000\exhyphenpenalty=50\arraybackslash}X >{\hsize=0.681\hsize\RaggedRight\hyphenpenalty=10000\exhyphenpenalty=50\arraybackslash}X}
\toprule
Configuration & Train items & Epochs & T3 BalAcc (five classes) & Eval loss \\
\midrule
T2-only & 6,949 & 3 & 0.757 & — \\
T3-only & 25,460 & 3 & 0.796 & — \\
Mixed & 32,409 & 3 & 0.882 & 0.035 \\
Mixed & 32,409 & 10 & 0.781 & 0.099 \\
\bottomrule
\end{tabularx}
\end{table}

The mixed corpus exceeds the T3-only specialist by 8.6 points and the T2-only specialist by 12.5 points on this diagnostic. Extending training from three to ten epochs reduces it by 10.1 points while evaluation loss rises from 0.035 to 0.099, which is the over-fitting signal discussed in Section 5.4.

\end{document}